\documentclass{article}

    \PassOptionsToPackage{numbers, compress}{natbib}

 \usepackage[main, final]{neurips_2024}

\usepackage{graphicx}
\usepackage{makecell}
\usepackage[utf8]{inputenc} 
\usepackage[T1]{fontenc}    
\usepackage{hyperref}       
\usepackage{url}            
\usepackage{booktabs}       
\usepackage{amsfonts}       
\usepackage{nicefrac}       
\usepackage{microtype}      
\usepackage{enumitem}
\usepackage{multicol}
\usepackage{hhline}
\usepackage{kotex}
\usepackage[capitalize]{cleveref}
\usepackage{multirow}  
\usepackage[table]{xcolor}

\title{\includegraphics[height=1.5em]{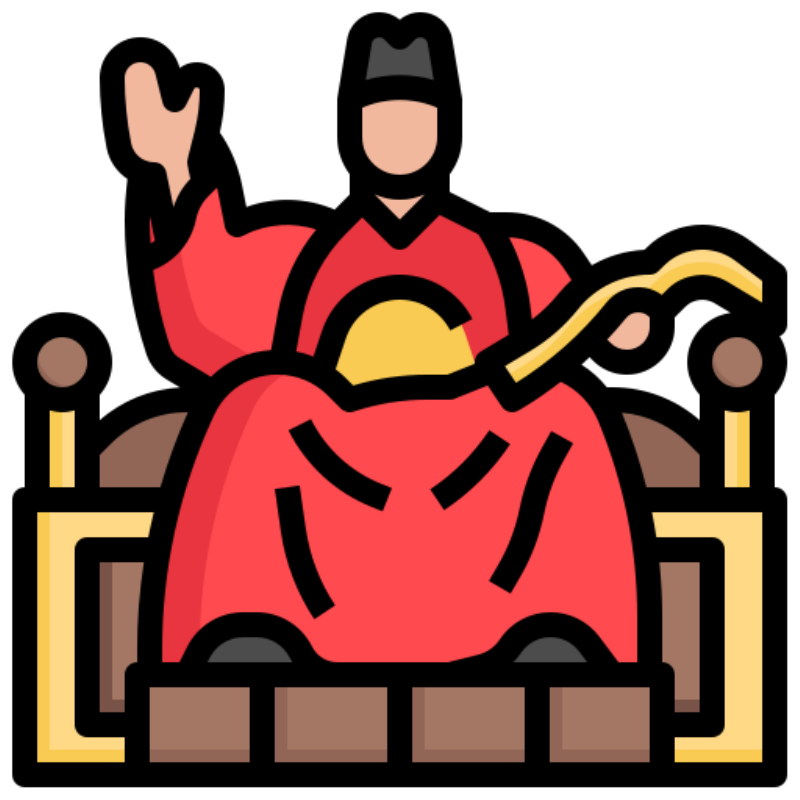}
Mi:dm 2.0\\ Korea-centric Bilingual Language Models}

\author{Tech. Innovation Group, KT \\
        midm-llm@kt.com \\
        \includegraphics[height=1em]{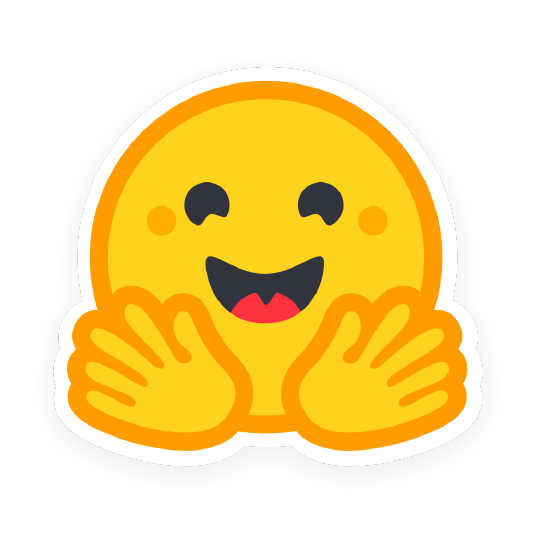}
        \url{https://huggingface.co/K-intelligence}}

\begin{document}
\maketitle
\begin{abstract}
We introduce Mi:dm 2.0, a bilingual large language model (LLM) specifically engineered to advance \textbf{\textsc{Korea-centric AI}}. This model goes beyond Korean text processing by integrating the values, reasoning patterns, and commonsense knowledge inherent to Korean society, enabling nuanced understanding of cultural contexts, emotional subtleties, and real-world scenarios to generate reliable, culturally appropriate responses.
To address limitations of existing LLMs—often caused by insufficient or low-quality Korean data and lack of cultural alignment—Mi:dm 2.0 emphasizes robust data quality through a comprehensive pipeline that includes proprietary data cleansing, high-quality synthetic data generation, strategic data mixing with curriculum learning, and a custom Korean-optimized tokenizer to improve efficiency and coverage.
To realize this vision, we offer two complementary configurations: \textbf{\textit{Mi:dm 2.0 Base}} (11.5B parameters), built with a Depth-up Scaling strategy for general-purpose use, and \textbf{\textit{Mi:dm 2.0 Mini}} (2.3B parameters), optimized for resource-constrained environments and specialized tasks. Mi:dm 2.0 achieves state-of-the-art performance in Korean-specific benchmarks, with top-tier zero-shot results on KMMLU and strong results in internal evaluations across language, humanities, and social science tasks.

The Mi:dm 2.0 lineup is released under the MIT license supporting extensive research and commercial use. By offering these accessible and high-performance Korea-centric LLMs, KT aims to accelerate AI adoption across Korean industries, public services, and education, while strengthening the Korean AI developer community and laying the groundwork for the broader vision of K-intelligence. Our models are available via \url{https://huggingface.co/K-intelligence}. For technical inquiries, please contact \url{midm-llm@kt.com}.

\end{abstract}

\section{Introduction}
\label{section1:introduction}
\subsection{The Korea-centric AI, Mi:dm 2.0}
KT (Korea Telecom) has developed Mi:dm 2.0 as an instruction-tuned language model that embodies what we call \textbf{\textsc{Korea-centric AI}}. Korea-centric AI refers to a model that thoroughly internalizes the unique values, cognitive frameworks, and commonsense reasoning intrinsic to Korean society. It is not simply about processing and responding in Korean; it is about the profound understanding that reflects and respects the socio-cultural fabric of Korean norms and values.

The development of Mi:dm 2.0 begins with a critical observation: despite the release of numerous large language models (LLMs) supporting the Korean language, few are truly grounded in the realities of Korean society. We identify a pervasive problem where existing LLMs, often trained on insufficient or low-quality Korean datasets, exhibit limited linguistic performance and a noticeable gap in their alignment with Korean cultural sensibilities \cite{kim2024open, kim2025thunderllmefficientlyadaptingllms}. This problem often leads to unnatural or emotionally incongruent responses, or even responses in languages other than Korean, from the perspective of Korean users. To directly address these deficiencies, we conceive Mi:dm 2.0, aiming to establish a new standard for a genuine Korea-centric AI that deeply internalizes the unique values, cognitive frameworks, and common sense reasoning inherent to Korean society, moving beyond mere linguistic proficiency to embody cultural nuance.

Our journey begins with a robust data curation pipeline, which meticulously defines criteria for sourcing high-quality, culturally representative Korean text, complemented by innovative synthetic data generation techniques. Following this, we detail our sophisticated pre-training methodology, which ensures effective learning even with a reduced corpus size through careful data selection and strategic training. We then delve into our model optimization techniques, designed to deliver computational efficiency that surpasses comparable domestic and international models. Furthermore, we outline the post-training techniques employed to significantly enhance the model's ability to perform Korean socio-cultural reasoning and generate contextually appropriate responses. Finally, we provide comprehensive quantitative and qualitative evaluations of Mi:dm 2.0's performance, including benchmark comparisons that unequivocally demonstrate its advanced understanding of Korean language, culture, and society.


\subsection{Mi:dm 2.0 Line-up: Base \& Mini}
We build Mi:dm 2.0 of two distinct parameter scales—11.5B and 2.3B—to meet diverse deployment needs. The Mi:dm 2.0 lineup is the result of extensive and systematic experimentation across a wide range of model architectures, parameter scales, and compression techniques. Through rigorous testing and iterative refinement, we identify configurations that deliver strong quantitative benchmark performance, robust language model evaluations, and reliable results from comprehensive human assessments using real-world scenario prompts.

Mi:dm 2.0 Base (11.5B) serves as a general-purpose foundation model, meticulously engineered to strike an optimal balance between scale and performance. Its development began with training an 8B-parameter model from scratch using KT's proprietary pre-training corpus. To further enhance its capabilities and reach the 11.5B scale, we then applied a Depth-up Scaling (DuS) strategy~\cite{kim2023solar}. This innovative approach efficiently expands the model's depth without requiring complex architectural changes, enabling us to effectively leverage representations learned by the initial 8B model.

In contrast, Mi:dm 2.0 Mini (2.3B) provides a lighter and more compact alternative. It is specifically optimized for deployment on resource-constrained devices, prioritizing computational efficiency. Mi:dm 2.0 Mini emphasizes task specialization, with a particular focus on intent understanding and machine translation, making it highly efficient for specific applications where resources are limited. \cref{tab:midm_model_specs} provides detailed configurations for both models.

During training, we employ various optimization techniques, including parallelization and quantization, to maximize GPU resource efficiency. We also extend and customize the underlying training framework to meet Mi:dm’s specific requirements. Mi:dm 2.0 delivers competitive or superior quality compared to open-source models of similar or larger scale while maintaining significantly lower computational overhead. We provide more detailed descriptions of model design and training methodology in ~\cref{section3:pre-training}.

Both Mi:dm 2.0 Base and Mi:dm 2.0 Mini are released under the MIT license, allowing extensive use for both research and commercial purposes. The models are available for download on Hugging Face and can be applied to a wide range of applications without restriction.


Looking ahead, we plan to expand the Mi:dm 2.0 lineup to support an even broader range of use cases, further solidifying its role as a foundation of K-\textit{intelligence.} We are also committed to releasing our training code, serving environments, and other research artifacts to facilitate broader adoption and continuous improvement within the ecosystem. Through these efforts, we aim to accelerate AI transformation (AX) across Korean industry, public services, and education, and to make meaningful contributions to the Korean AI developer community with both the Mi:dm 2.0 Base and Mi:dm 2.0 Mini.

\begin{table}[ht]
\centering
\renewcommand{\arraystretch}{1}
\footnotesize
\begin{tabular}{l| c| c| c| c}
\toprule
\textbf{Models} & \textbf{Model Type} & \textbf{Model Size} & \textbf{Context Length} & \textbf{License} \\ 
\midrule
Midm 2.0 Base & Dense & 11.5B & 32K & MIT \\
Midm 2.0 Mini & Dense & 2.3B & 32K & MIT \\
\bottomrule
\end{tabular}
\vspace{6pt}
\caption{Mi:dm 2.0 line-up architecture and license}
\label{tab:midm_model_specs}
\end{table}

\section{Data}
\label{section2:data}
This section details the methodologies for data construction and management, including domain classification, document filtering pipeline, and synthetic data generation strategies adopted in the development of Mi:dm 2.0. Recent LLMs are known to critically depend on large amounts of high-quality textual data to achieve robust performance~\cite{chang2024scaling,iskander2024quality}. However, the Korean language poses a unique challenge due to the limited availability of high-quality and publicly accessible training data compared to English. Furthermore, the heterogeneous quality of Korean corpora available hinders stable and reliable performance.~\cite{son2023haerae,shin2024kulture}
 
To overcome the structural limitations of existing Korean training datasets, Mi:dm 2.0 is strategically designed and robustly trained from the earliest stages of pre-training corpus construction. We explicitly prioritize data quality over quantity to ensure the stable assimilation of both general knowledge and the nuanced Korean cultural and societal context. Despite the reduction in overall token volume, we focus on filtering for accurate, complete, and trustworthy documents. 

For Mi:dm 2.0, we define high-quality data as documents that are contextually coherent, highly readable, non-toxic, and well-formed. To achieve this standard, our proprietary data cleansing pipeline strictly excludes documents that do not meet these criteria.
Unlike English datasets, where quality selection is simpler due to the abundance of commercially available open-source corpora, applying the same stringent quality criteria to Korean text significantly reduces the total usable tokens. Nevertheless, we firmly maintain our focus on this data quality approach to enhance the model's overall representational power and generalization performance.

To address this, we deliberately generate high-fidelity synthetic data to supplement the limited volume of high-quality Korean corpora. This synthetic data, primarily grounded in organic (human-generated) content, is augmented by language models. Given the overarching challenge of limited availability and high acquisition costs for Korean datasets, many of which are web-based, synthetic data augmentation has proven to be particularly valuable. These include translating English corpora into Korean and generating textbook-style documents from topics and keywords extracted from existing bilingual corpora.

Furthermore, Mi:dm 2.0's data engineering pipeline incorporates data mixing and curriculum learning strategies. Concurrent with corpus filtering and augmentation, we establish a hierarchical domain taxonomy based on application-specific needs. This taxonomy then guides the alignment of training data distribution with the intended use cases of the model. To systematically monitor and manage dataset balance, we train a domain classifier across the entire corpus to quantify its distribution. This enables the model to identify underrepresented domains quantitatively and granularly. For domains lacking sufficient token density, we generate additional synthetic data via a feedback loop, enhancing both coverage and diversity of the training corpus.

To optimize data efficiency, our model utilizes a custom tokenizer specifically designed to capture the unique linguistic characteristics of Korean. Based on a precisely curated pre-training corpus, we design the Mi:dm 2.0 tokenizer to handle the morphological structure of the Korean language more effectively than traditional GPT-series tokenizers. This approach achieves higher token compression and significantly enhances computational efficiency during both training and inference.

The entire data pipeline for our model adheres to rigorous standards for data provenance, licensing, and compliance. All training data are sourced from open-source datasets or formal licensing agreements with third parties, ensuring legitimacy. We strictly exclude any data posing legal or ethical concerns, including unauthorized crawls or user-sensitive content. Notably, we exclude Personally Identifiable Information (PII) and proprietary customer data, thereby ensuring both data security and ethical responsibility throughout the model development process.
\begin{table}[ht]
\centering
\renewcommand{\arraystretch}{0.85}
\resizebox{0.4\linewidth}{!}{\tiny
\begin{tabular}{ll}
\toprule
\textbf{Category} & \textbf{Details} \\ 
\midrule
\textbf{Language} & English \\
                  & Korean \\
                  & Code \\
                  & Math \\
                  & Multi Language \\
\midrule
\textbf{Source} & Organic \\
                & \quad\quad Web \\
                & \quad\quad Government \\
                & \quad\quad Book \\
                & \quad\quad News \\
                & \quad\quad Paper \\
                & \quad\quad Encyclopedia \\
                & \quad\quad Others \\
                & Synthetic \\
\midrule
\textbf{Domain} & Humanity \\
                & STEM \\
                & Applied Science \\
                & Health \& Food \\
                & Life \& Culture \\
                & ETC \\
\midrule
\textbf{Expression Mode} & Written \\
                         & Spoken \\
\midrule
\textbf{Stylistic Tone} & Formal \\
                        & Informal \\
\bottomrule
\end{tabular}
}
\vspace{6pt}
\caption{Data Categorization: Categories and Details}
\label{tab1:data_categorization}
\end{table}

\subsection{Multidimensional and Hierarchical Data Classification Framework}
\label{subsection2-1:data}
Precise analysis and structured management of training data are essential for developing high-performance language models. However, the scarcity of detailed categorization for Korean language datasets in existing research significantly limits efforts to expand data coverage and interpret model performance~\cite{kim2024click}.
To overcome these challenges, we define a novel data classification framework designed to support balanced data distribution and efficient training. It organizes data across multiple dimensions -including language, domain, source, and linguistic style -and is consistently applied throughout the entire data pipeline, from collection to training, as shown in ~\cref{tab1:data_categorization}. Hence, following the refinement and high-quality data selection, we classify data from multiple perspectives, as detailed below.

In detail, from the language perspective, our dataset is classified not only as multilingual text, such as Korean and English, but also as non-linguistic content, including mathematical expressions and source code. From a domain perspective, our dataset is categorized using an internally developed taxonomy designed to reflect both the thematic content and the intended application of each document. This taxonomy comprises six primary domains: Humanity, STEM, Applied Science, Health \& Food, Life \& Culture, and ETC, along with 20 mid-level subdomains that provide further granularity. Finally, for the data source, documents are broadly categorized as either organic or synthetic. Organic data consists of naturally occurring text derived from real-world human activity. This includes sources such as web pages, news articles, books, encyclopedias, government documents, academic papers, and other written materials. Conversely, synthetic data refers to text generated with augmentation techniques. This includes documents created through machine translation, document rewriting, and advanced methods such as Chain-of-Thought (CoT) generation~\cite{wei2022chain}. Beyond these dimensions, we also classify texts according to their linguistic style and tone. Specifically, we categorize documents based on whether they predominantly feature written or spoken language characteristics and whether their tone is formal or informal.

Throughout the entire Mi:dm 2.0 training process, statistical information for each classification attribute is continuously managed across all training data subsets, from data collection to training, guided by this classification framework. Each training sample is tagged with up to five classification attributes in these dimensions. For instance, we would categorize a Korean-language web review of a children's book titled after the historical figure "King Sejong (세종대왕, Sejong Daewang)" as "Korean" in the language dimension, as "History" subdomain within the "Humanity" domain, and as "Organic" in source. In another case, a Korean document generated by extracting the keyword "chlorophyll" from an English web article, rewriting the content in a textbook-like format, and translating it into Korean would be classified as "Korean" in language, "Biology" subdomain within the "STEM" domain, and as "Synthetic" in source. By maintaining comprehensive statistics across all classification axes, we can monitor data distribution, identify underrepresented categories, and strategically augment the dataset to ensure balance and diversity.


\subsection{Composition of Data Sources}
\label{subsection2-2:data}
Among the multiple classification axes described in ~\cref{subsection2-1:data}, data source serves as a critical criterion from the earliest stages of data collection. The goal is to ensure diversity within the pre-training corpus, enabling the language model to acquire expressive capabilities in various document styles and topics. This ultimately enhances its ability to generalize linguistically across different real-world contexts.

\begin{figure}[htbp]
  \centering
  \includegraphics[width=0.8\linewidth]{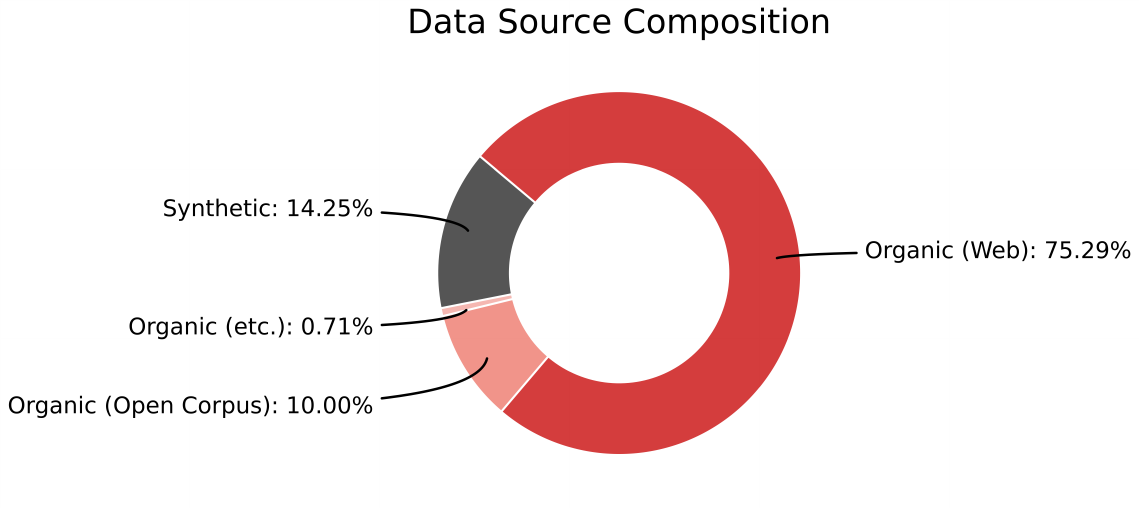}  
  \caption{Distribution of Dataset by Source Type (Organic vs. Synthetic)}
  \label{fig1:figure2}
\end{figure}

Accordingly, the pre-training corpus primarily consists of organic data —naturally occurring, human-authored text—to accurately reflect authentic language usage environments, as shown in ~\cref{fig1:figure2}. Approximately 85.7\% of the total dataset is sourced from organic domains, such as web documents. This composition is carefully selected to enable the model to acquire high-fidelity knowledge of Korean-specific syntactic structures, discourse patterns, and informal expressions. Additionally, approximately 10\% of the corpus comprises of open-source, organic data acquired from high-quality public datasets, such as AIHub\footnote{\url{https://www.aihub.or.kr}} and the National Institute of the Korean Language (NIKL)\footnote{\url{https://www.korean.go.kr}}. These resources are comprised of administrative documents, transcribed spoken dialogues, and other publicly available linguistic assets, thereby enhancing the model's reliability, particularly in terms of standard language usage and compatibility with public benchmarks. A smaller subset—approximately 0.71\%—is comprised of additional organic sources, including academic papers, books, government documents, and dictionaries. Although this category constitutes a relatively minor portion of the corpus, it significantly contributes to the performance of the model by expanding its capacity to understand high-density informational text, formal language, and conceptually coherent content. This data composition strategy extends beyond exclusively ensuring topical variety. It aims to provide the model with a broad range of linguistic contexts found in real-world Korean use, helping it become more natural, robust, and contextually aware.

To enable our model to effectively embody `Korea-centric AI,' the data acquisition and corpus construction strategy prioritized critically maximizing the linguistic and topical diversity of its Korean-language organic dataset. We employ a systematic corpus construction methodology that integrates both public collections and licensed acquisitions. This approach yields a comprehensive array of resources, including Korean literary works, modern historical records (e.g., news articles), public documents (e.g., legal texts and dictionaries), and structured databases specifically curated for Korean cultural heritage.

The major sources of organic data derive from Common Crawl (CC)\footnote{\url{https://commoncrawl.org}}, Hugging Face\footnote{\url{https://huggingface.co}}, AIHub, and the NIKL. News articles, books, and dictionaries are obtained via formal licensing agreements and subsequently filtered according to internal quality standards. For the English web corpus, we leverage open-source corpora with pre-annotated document-level quality indicators. In contrast, Korean web data is sourced from CC-based corpora and processed through an internally developed filtering pipeline to ensure suitability. Furthermore, data acquired from AIHub and NIKL undergo explicit permissioning and rigorous curation to ascertain their applicability for commercial deployment.

Synthetic data accounts for approximately 14\% of the entire training dataset. Its primary purpose is to compensate for domain-specific data scarcity in organic Korean datasets, specifically to augment coverage in underrepresented areas. Furthermore, synthetic data generation was employed to enhance knowledge about Korea that was insufficient in the original organic data source. For English, given the abundance of high-quality open datasets, we selectively integrate publicly available synthetic corpora. Conversely, we produce Korean synthetic data via custom-built generation pipelines. The construction of synthetic data in our model leverages both publicly available research methodologies and proprietary augmentation techniques developed in-house~\cite{abdin2024phi,chen2024diversity}. These synthetic data generation strategies are further elaborated in Section 2.4.

\subsection{High-Quality Data Filtering and Refinement Pipeline}
Mi:dm 2.0 implements an intentionally developed quality control pipeline for data selection and pre-processing, ensuring that the resulting corpus is optimally suited for next-token prediction training. The filtering criteria are defined to construct token sequences that are both coherent and learnable, minimizing interference during model training.

From a pre-training perspective, high-quality data is defined as text that meets the following conditions:
\begin{enumerate}[label=\arabic*), leftmargin=2em, itemsep=1ex]
  \item Data should maintain consistent textual coherence, devoid of disruptive special characters or grammatical malformations.
  \item Data that satisfies 1) and should be comprised of highly readable and well-formed complete sentences, adhering to standards of linguistic completeness.
  \item Data should be devoid of harmful content and free from any personally identifiable information that could compromise privacy.
\end{enumerate}


These quality standards are applied throughout the data preparation pipeline, particularly during the pre-training phase, where the generalization capability of the model is shaped. By minimizing the incidence of noisy or irrelevant tokens, the model can learn more stable token distributions and more effectively internalize linguistic patterns and knowledge.

To achieve this, first our model employs multi-stage data filtering and refinement pipeline that is both language- and source-specific, with a particular emphasis on Korean-language data to mitigate the relative scarcity of high-quality Korean tokens.
\begin{figure}[htbp]
  \centering
  \includegraphics[width=0.82\linewidth]{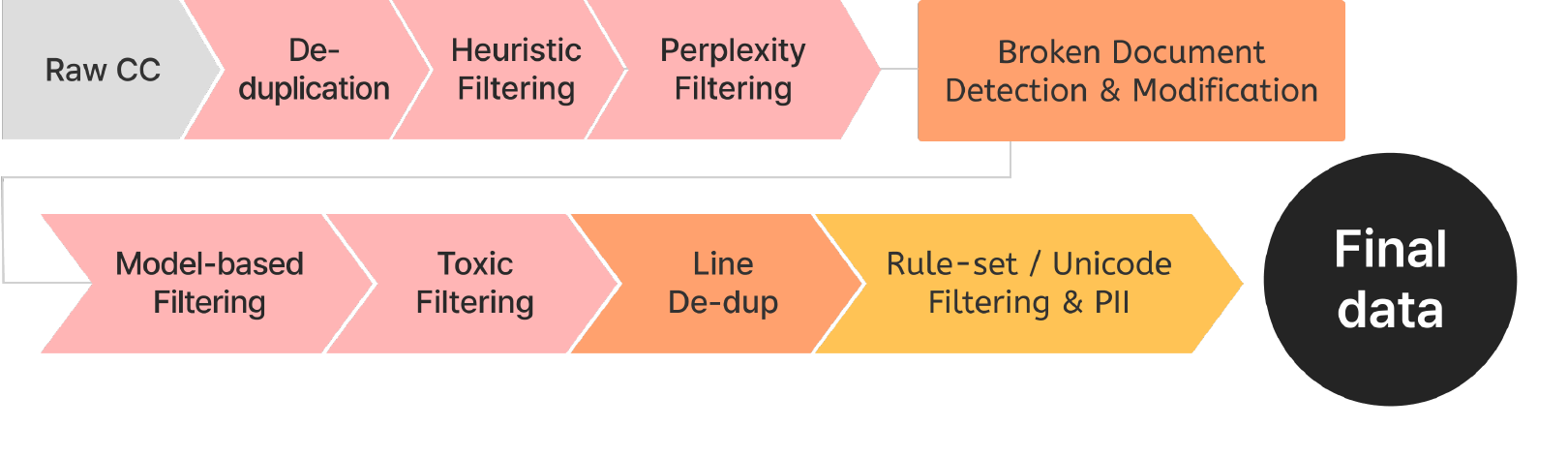}  
  \caption{Multi-Stage Filtering and Refinement Pipeline for Korean Web Documents}
  \label{fig2:figure3}
\end{figure}
As illustrated in ~\cref{fig2:figure3}, the primary component of this effort is an internally developed, 8-stage filtering pipeline for large-scale Korean web data. Web corpora—especially those derived from CC —contain documents with format corruption, low-quality text, harmful or biased content, or PII, rendering raw web data unsuitable for direct use. Our model addresses this by applying a rigorous, sequential filtering strategy specifically tailored to Korean web content. Each stage of the pipeline incrementally refines the dataset, isolating documents of genuine learning value:

\begin{enumerate}[label=\arabic*), leftmargin=2em, itemsep=1.2ex]
\item \textbf{Document De-duplication}: Redundant documents are removed based on cosine similarity over TF-IDF vectors.
\item \textbf{Heuristic Filtering}: Documents containing hashtags, excessive ellipses, or abnormal punctuation are filtered using handcrafted rules inspired by prior work~\cite{kim2023solar}.
\item \textbf{Perplexity Filtering}: Documents exhibiting abnormal n-gram perplexity are removed as likely being low-quality or incoherent.
\item \textbf{Broken Document Detection and Correction}: Unicode corruption and broken character sequences are detected and corrected.
\item \textbf{Model-Based Quality Filtering}: An ensemble of binary classifiers is trained using annotated examples of high- and low-quality documents. An ensemble of binary classifiers is used, comprising one classifier trained on general quality criteria and another trained on educational quality criteria inspired by prior studies~\cite{penedo2024fineweb, su2024nemotron}.
\item \textbf{Toxic Content Filtering}: A binary classifier trained on KT’s proprietary Korean toxicity and bias taxonomy is used to eliminate harmful or offensive content.
\item \textbf{Line-Level De-duplication}: Within a document, repetitive lines or paragraphs are removed to reduce redundancy.
\item \textbf{Final Rule-Based Refinement and PII Anonymization}: Final cleanup includes Korean-specific formatting fixes, normalization of invisible Unicode tokens, and removal or anonymization of any detected personal information.
\end{enumerate}

For the subsequent step, we design a source-specific refinement pipeline for non-common Crawl Korean datasets, such as those securely acquired from books, encyclopedias, academic papers, expert knowledge databases, and licensed news articles. This pipeline consists of source-specific refinement modules for each domain and explicitly reflects the unique characteristics of Korean language data. 


For instance, the news article refinement module incorporates rules for removing strings that are irrelevant to the core content of individual articles or disrupt the overall context. The bylines (reporter names and email addresses) at the end of Korean articles or image captions remaining within the main body after image removal are removed. Additionally, rules applicable to Korean data, such as eliminating string patterns like `[속보]' or `상보', which are uniquely found in domestic Korean online news headlines, are also incorporated.

As another example, the Korean court judgment refinement module applies rules to redact PII and extract only the critical content from the unique formatting of domestic court judgments while preserving the structural and semantic content necessary for learning formal legal language.

Documents refined through these source-specific rules are considered comparable in quality to those that have passed harmful content filtering in the Korean web data pipeline. Subsequently, they undergo final steps such as deduplication and PII anonymization before being used in the training dataset.

Finally, for English, code, and mathematical content, high-quality, commercially suitable public datasets are selected and rigorously assessed during acquisition through manual inspection and sample-based qualitative analysis. Deduplication and final normalization are performed to ensure alignment with the quality standards of the Korean dataset.

\subsection{Synthetic Data Generation}
It is essential to acquire high-quality data, encompassing diverse knowledge and linguistic expressions across a wide range of domains, for our model to enable complex reasoning and conceptual understanding. In particular, being Korea-centric AI requires gathering a collection of Korean-language datasets that reflect sufficient diversity and representativeness. However, in practice, the availability of open-access Korean-language corpora is significantly lower than that of English, and a large portion of the available data is concentrated in web sources, which often contain low-quality content. Furthermore, the majority of Korean-language data is disproportionately biased towards humanities and social sciences, creating additional challenges for domain diversity.

\begin{figure}[htbp]
  \centering
  \includegraphics[width=0.8\linewidth]{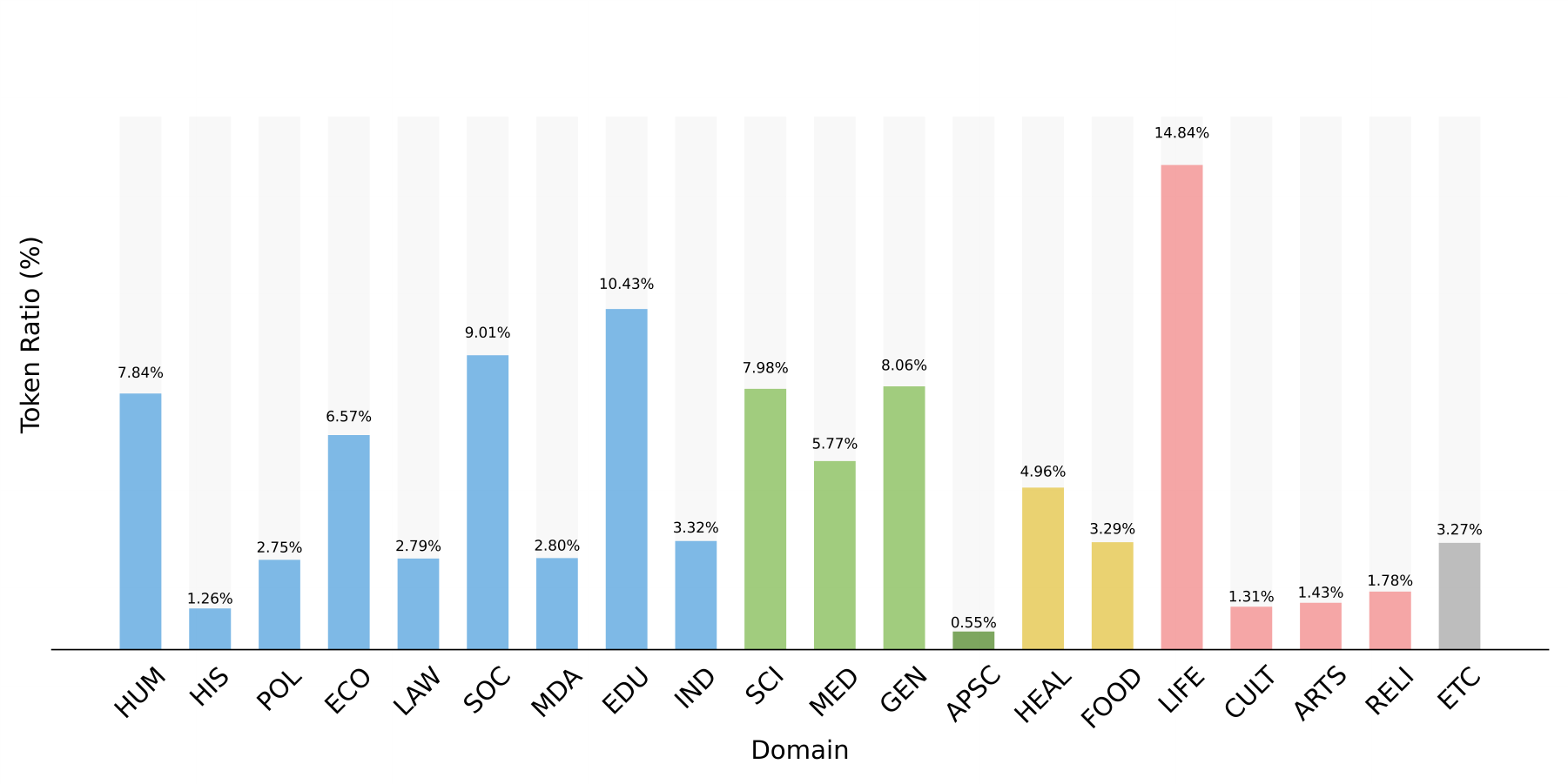}  
  \caption{Domain-wise token distribution of Mi:dm 2.0 pretraining corpus}
  \label{fig3:figure4}
\end{figure}

These structural limitations are empirically observed in the data distribution statistics collected after sourcing from various domains. ~\cref{fig3:figure4} illustrates the token distribution of Mi:dm 2.0’s pretraining corpus across data sources and domains. The light blue bars, which represent tokens from the humanities and social sciences, account for a disproportionately large portion of the dataset. In contrast, domains such as applied sciences (APSC), arts (ARTS), and culture (CULT) are severely underrepresented. Notably, applied sciences constitute only 0.1\% of the total token count, highlighting a clear example of domain imbalance within the current data ecosystem. Such an imbalance has the potential to affect the model’s expressiveness and reasoning capabilities in domain-specific applications.

To mitigate such biases, Mi:dm 2.0 strategically incorporates high-quality synthetic data during the pre-training phase. The generation pipeline extends simple translation-based augmentation, specifically designed to simulate reasoning structures and compositional understanding by producing textbook-style explanatory passages, logically structured reasoning chains, and diverse document types tailored to specific learning objectives. All synthetic data undergoes the same rigorous post-processing as the organic corpus, including Unicode normalization, PII filtering, and deduplication, to ensure consistency in quality before being included in the final training dataset.

\subsubsection{Corpus Construction for Domain Balancing in Korean-Language Data}
An analysis of Mi:dm 2.0’s pre-training data distribution reveals that domains, particularly those underrepresented in the Korean corpus collected from public datasets, are computationally intensive fields such as STEM (science, technology, engineering, and mathematics) and economics. Such imbalance in the dataset is reflected in early benchmark evaluations during model development, where our model consistently underperformed in disciplines that demand high-level reasoning and domain-specific knowledge, such as physics, chemistry, biology, mathematics, computer science, and economics.

To address this structural bias, we systemically collect high-reliability open-source materials to serve as seed data for domain-targeted synthetic augmentation. These seed corpora are carefully selected to introduce and reinforce conceptual understanding and problem-solving capabilities in previously underrepresented domains. Leveraging insights from prior work~\cite{chen2024diversity,abdin2024phi}, we tailor prompt structures for each domain. Specifically, core concepts derived from seed documents are converted into high-quality Korean language instructional texts, utilizing various formats, including textbook-style explanations and scenario-driven narratives. The synthetic documents are generated at varying levels of difficulty and designed for diverse reader profiles, thereby enriching the corpus in domains lacking sufficient coverage.

\subsubsection{Reconstruction and Augmentation of Non-Selected Korean Web Documents}
The portion of Korean language in the CC corpus integrated into our model's pre-training dataset stems from a widely used open-source resource for large-scale language model training. Despite applying a rigorous filtering pipeline to extract only a small subset of high-quality documents, the CC corpus still holds the largest portion of the Korean-language dataset used in our model. 

However, this dataset is inherently noisy, with a significant portion of its documents being of low quality. In fact, over 80\% of the initially collected raw CC data is subsequently excluded during the data filtering process. This highlights a structural limitation of CC: while it provides broad coverage, the usable token yield relative to curation effort remains low.

To overcome this inefficiency and increase the number of usable Korean-language tokens for pre-training, we develop a rewriting-based synthetic reconstruction strategy for a portion of the filtered-out CC documents. Manual inspection of rejected samples reveals that some documents—although initially discarded due to poor formatting or content—can be transformed into high-quality training material if their core topics and sentence structures are properly reorganized. Given the lack of consistent structural patterns, these documents cannot be effectively recovered using rule-based methods. Consequently, we developed a generative rewriting pipeline specifically for Korean CC documents.

This rewriting process is composed of two prompt stages. In the first stage, the topic analysis module extracts metadata, including the central topic of the document and relevant paragraph indices. This enables the filtering of irrelevant fragments within documents—such as image captions, template-based advertisements, or copyright notices—often found as short, extraneous sentences. This stage also identifies and separates cases where a single document actually contains multiple unrelated articles concatenated together. In the second stage, based on the topic structure extracted earlier, the pipeline generates excerpted and rewritten documents, focusing only on topic-relevant content. The rewriting model synthesizes new documents that preserve the central meaning of the original while eliminating noise and improving coherence.

Finally, all reconstructed documents are passed through the same web corpus filtering pipeline used for the original CC documents. This ensures that any rewritten documents containing harmful content, biased language, or incoherent structure are excluded. Only documents that met the same high-quality criteria as the original filtered set are included in the final pre-training corpus.

\subsubsection{Structural Augmentation of English Web Documents}
Our strategy synthetically enhances the structural diversity and complexity of Korean language web data, which is primarily derived from CC. Despite CC’s known limitations—namely, its high proportion of low-quality content—it remains a valuable corpus as it closely reflects modern language use in real-world human contexts. 

In English-speaking communities, several high-quality web corpora curated from CC have been released, accompanied by research on large-scale web data filtering methods~\cite{su2024nemotron,li2024datacomp}. These efforts have provided models with rich, well-structured input across various formats.

In contrast, the Korean CC corpus tends to be limited in both structural diversity and topical breadth, being heavily skewed toward specific formats such as news articles, blogs, and online community posts. As a result, it lacks the structural richness and stylistic complexity observed in English corpora. Notably, Korean CC contains relatively few examples of long-form structured documents or intent-driven formats (e.g., summarization, QA, translation), which are indispensable for training models on higher-order reasoning and downstream tasks such as question answering.

To mitigate these limitations, our model integrates a cross-lingual synthetic augmentation strategy, where unused English web samples are rewritten into Korean texts during pre-training. These rewritten documents are not direct translations, but content-preserving rewrites into natural Korean formats that differed structurally from the original web style. This approach helps avoid typical translation errors—such as the literal rendering of idiomatic expressions or the misinterpretation of domain-specific terminology—often seen in naïve machine translation.

For example, the content of an English web document can be transformed into the style of a Korean university entrance exam question in the "Speaking and Writing" section. In this process, the original content is transformed into a coherent Korean passage, accompanied by reading comprehension questions and answer sets. This enables the model to learn both richly composed texts and QA-style supervision. In practice, the majority of synthetic QA documents contributed solely the passage portion to the final pre-training corpus, while a select subset retained the QA pairs. The QA dataset included in the corpus is further filtered through CoT-based verification, and only samples with verified correct answers are retained.

Meanwhile, we opt not to integrate any additional rewriting on previously synthesized data to circumvent the risks of semantic drift, factual inconsistency, or unintended duplication. This decision preserves the integrity and quality of the final corpus.

\subsubsection{Structured Long Chain-of-Thought Data for Math and Code Reasoning}
We construct the LongCoT dataset~\cite{wei2022chain} to provide synthetic problem-solving sequences that explicitly model multi-step reasoning. Each math or code example includes a clear, logically structured solution path designed to help the model learn the reasoning patterns required for complex tasks.

This resource not only supplements the limited availability of Korean-language data but also improves the model’s ability to learn reasoning in Korean. All solutions and explanations are written primarily in Korean to help the model develop native-level logical reasoning for structured problem domains.

The final data is formatted into pre-training-ready text segments and integrated directly into the Mi:dm 2.0 training set. By exposing the model to high-quality reasoning demonstrations early on, this approach supports stronger performance in math, programming, and structured question-answering tasks.

\section{Pre-training}
\label{section3:pre-training}
This section outlines the model lineup of Mi:dm 2.0 and the corresponding pre-training strategies applied to each model. Mi:dm 2.0 achieves efficient pre-training and strong performance in Korean understanding and generation tasks by leveraging a small, highly curated Korean-language dataset with limited computational resources. This approach demonstrates the feasibility of developing competitive language models even in resource-constrained environments.

To introduce the pre-training methodology for each model in detail, ~\cref{section3.1} provides an overview of the model lineup expansion process. This is followed by subsections describing the pretraining techniques for each model variant: ~\cref{section3-1-1} covers Mi:dm 2.0 Base and ~\cref{section3-1-2} discusses Mi:dm 2.0 Mini. ~\cref{section3-1-3} then presents the sequence length extension techniques applied during training. Finally, \label{section3.2} details the computational cost optimization strategies that enable efficient large-scale training.

\begin{figure}[htbp]
  \centering
  \includegraphics[width=0.9\linewidth]{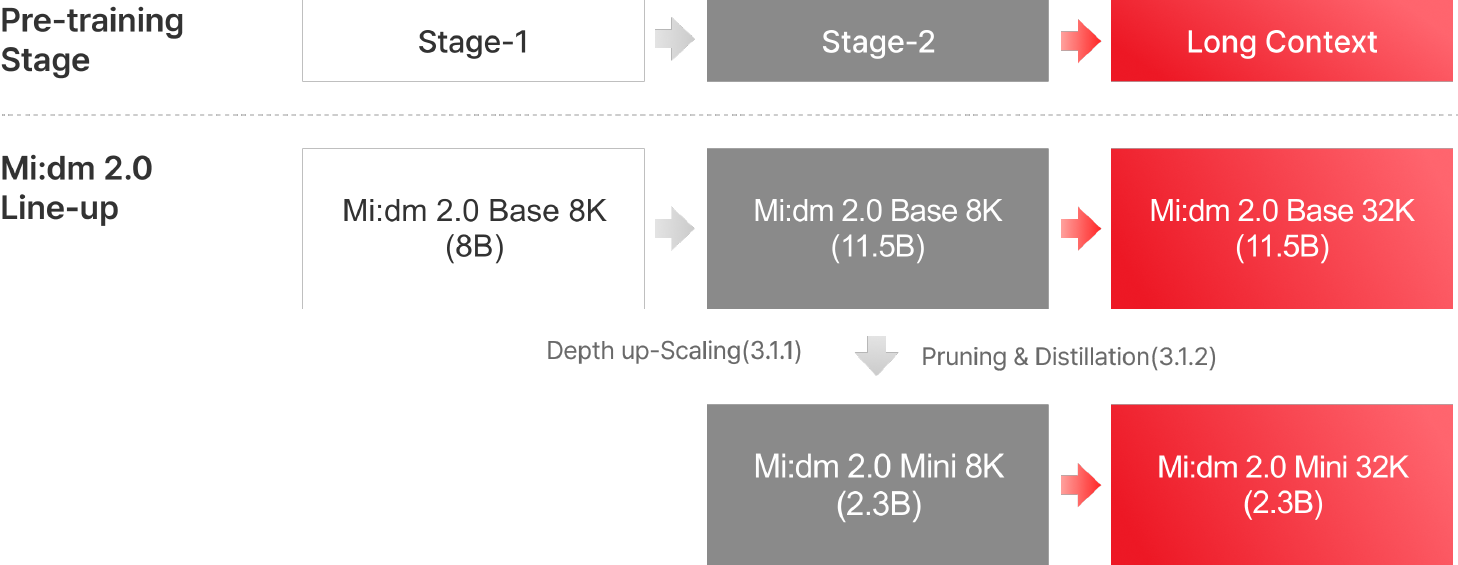}  
  \caption{Mi:dm 2.0 Model Lineup and Pre-training Pipeline}
  \label{fig4:figure5}
\end{figure}

\subsection{Model Architecture}
\label{section3.1} 
All Mi:dm 2.0 lineups are based on a Transformer decoder-only architecture ~\cite{grattafiori2024llama}. The number of layers in this architecture is adjusted to optimize training efficiency and performance across various model sizes and applications.

As illustrated in ~\cref{fig4:figure5}, the expansion of the model architecture employs a three-stage training procedure.
In the initial pre-training phase (Stage-1), an 8B parameters Mi:dm 2.0 Base is trained from scratch. This foundation model, entirely developed in-house from architectural design to training, does not leverage existing open-source weights. The Stage-1 model comprises of 32 layers of dense Transformer decoders and processes input sequences up to approximately 8,000 tokens. The objective of this stage is to establish a foundation model that can acquire general language abilities and extensive domain knowledge. To achieve this, the training dataset for Stage-1 is curated to cover a wide range of topics and domains.

In the second stage (Stage-2), the focus shifts to scaling up the model using the checkpoint from Stage-1. The outcome of this stage is the finalized Mi:dm 2.0 Base. Depth up-Scaling (DuS) technique~\cite{kim2023solar} is applied to increase the number of transformer decoder layers from 32 to 48, resulting in a parameter count of 11.5B. Although the volume of the training data in Stage-2 is smaller than that of Stage-1, it is composed of ultra-high-quality data to enhance the model's ability to generate refined, task-specific responses.

Based on the Mi:dm 2.0 Base, after completing its second training phase, Mi:dm 2.0 Mini is trained with a reduced size of 2.3B parameters through quantization. To preserve the core knowledge of the Mi:dm 2.0 Base while reducing the size, both width-based pruning techniques~\cite{mirzadeh2020improved, wang2020minilm} and multi-stage knowledge distillation~\cite{sreenivas2024llm, muralidharan2024compact} are employed. The resulting Mi:dm 2.0 Mini is optimized for on-device deployment, enabling efficient inference in resource-constrained environments.

In the final stage, long-context training is applied to both models. This involves extending the base frequency in Rotary Position Embedding (RoPE)~\cite{men2024base} to process longer input sequences. Whereas the original Mi:dm 2.0 Base and Mi:dm 2.0 Mini could only handle sequences up to approximately 8,000 tokens, long-context training extends their maximum input token length to approximately 32,000 tokens, allowing for more effective processing of long documents.
The details of each Mi:dm architecture are summarized in~\cref{tab3:midm_model_detail}.

\begin{table}[ht]
\centering
\footnotesize
\begin{tabular}{l|c|c}
\toprule
\textbf{Specification} & \textbf{Mi:dm 2.0 Mini} & \textbf{Mi:dm 2.0 Base} \\
\midrule
Number of Parameters & 2.3B & 11.5B \\
Hidden size & 1,792 & 4,096 \\
Number of layers & 48 & 48 \\
Activation function & SiLU & SiLU \\
Feedforward Dimension & 4,608 & 14,336 \\
Attention type & GQA & GQA \\
Number of attention heads & 32 & 32 \\
Head size & 128 & 128 \\
Context length & 32,768 & 32,768 \\
Positional Embeddings & RoPE($\theta$=8,000,000) & RoPE($\theta$=8,000,000) \\
Vocab size & 131,392 & 131,384 \\
Tied word embedding & True & False \\
\bottomrule
\end{tabular}
\vspace{6pt}
\caption{Detailed architectural specifications of the Mi:dm 2.0 models. 
Both variants share key design choices such as the number of layers, attention mechanism (GQA: Grouped Query Attention~\cite{ainslie2023gqa}), and activation function (SiLU~\cite{shazeer2020glu}), while differing in hidden size, parameter count, and feedforward dimensions. 
The table also reports positional embedding configuration (RoPE) and vocabulary size.}
\label{tab3:midm_model_detail}
\end{table}

\subsubsection{Mi:dm 2.0 Base: Depth Up-Scaling} 
\label{section3-1-1}
We design Mi:dm 2.0 Base for robust performance across diverse application environments, even with limited computational resources. This is achieved through DuS, a training methodology that systematically duplicates specific layers from the base model and stacks them on top of existing layers to increase the model's depth.

The efficacy of DuS depends on strategically choosing layers with strong representational capacity. Effectively choosing layers with strong representational capacity allows DuS to maximize its advantages, enhancing the expressiveness and performance of the model while efficiently reusing resources from the initial training phase.

For Mi:dm 2.0 Base, we adopt a quantitative methodology, as proposed in ~\cite{pengyarn}, to determine which layers to replicate. This approach measures the change in embedding representations before and after each layer by calculating the cosine similarity between them. Layers with higher cosine similarity scores are selected for duplication under the assumption that they stably preserve the input information with minimal degradation during the training.
\begin{figure}[htbp]
  \centering
  \includegraphics[width=0.87\linewidth]{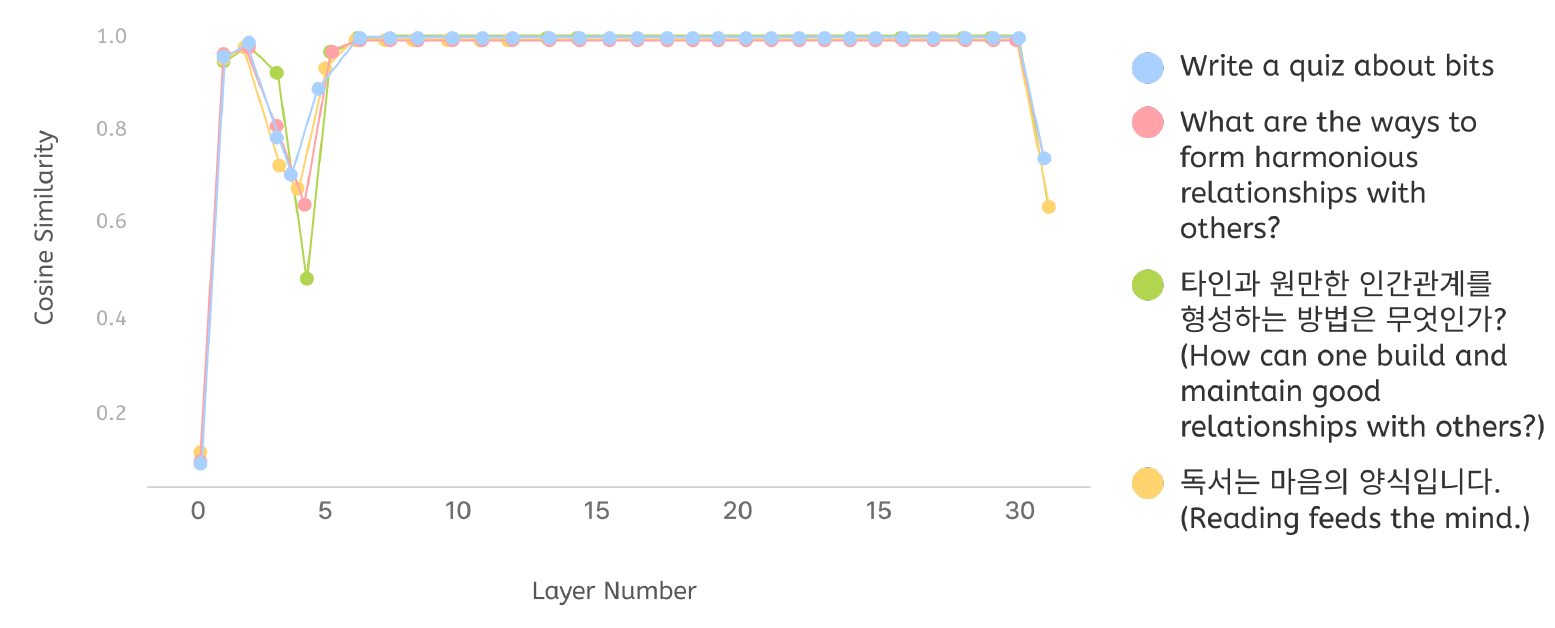}  
  \caption{Analysis of Layer-wise Embedding Changes Based on Cosine Similarity}
  \label{fig5:figure6}
\end{figure}

Applying this method, we analyze the embedding variation across all layers of the initial 8B-scale model used for Mi:dm 2.0 Base expansion. As shown in~\cref{fig5:figure6}, we compute the cosine similarity between pre- and post-layer embeddings for each token in the input sentence and then average the scores per layer. This analysis reveals that the embedding changes stabilize after the 5th layer, with layers between the 7th and 29th showing near-perfect similarity (close to 1.0), indicating minimal information loss.

We confirm these results through~\cref{fig5:figure6} and additional internal experiments on other language models with similar architectures, which show consistent patterns as reported in~\cite{pengyarn}. Based on these findings, we select the continuous range from the 7th to the 29th layers for duplication, extending the total number of layers to 48 in Mi:dm 2.0 Base.

To verify the effectiveness of this architectural expansion, we evaluate the performance of the expanded model in its initial state. As shown in~\cref{tab4:midm_performance}, the structure-extended but untrained Base-init model demonstrates comparable benchmark performance to the original Mi:dm 2.0 Base-8B, suggesting that our cosine similarity-based layer selection method successfully achieves effective structural scaling.

After this layer expansion step, Mi:dm 2.0 Base undergoes continual pre-training with thoroughly refined high-quality data to further improve its linguistic expressiveness and general domain understanding. This continual pre-training is structured in two stages: the first stage aims to ensure stable convergence immediately after expansion, while the second stage focuses on strengthening Korean and STEM domains using curated Korean data and custom-generated synthetic data. The training phase employs a Warmup-Stable-Decay scheduler~\cite{huminicpm} with a peak learning rate of 3e-4, decaying linearly to 0.0 during the final 10\% of training. As shown in ~\cref{tab4:midm_performance}, the results demonstrate improved performance across various domains, including Korean and mathematics.



\begin{table}[ht]
\centering
\footnotesize
\begin{tabular}{l|c|c|c|c|c}
\toprule
\textbf{Model} & \textbf{MMLU} & \textbf{HellaSwag} & \textbf{KMMLU} & \textbf{HAERAE} & \textbf{GSM8K} \\
\midrule
Mi:dm 2.0 Base-8B & 51.94 & 74.98 & 29.36 & 56.18 & 14.48 \\
Mi:dm 2.0 Base-init & 52.05 & 74.43 & 29.60 & 56.82 & 12.89 \\
Mi:dm 2.0 Base* & \textbf{62.61} & \textbf{79.36} & \textbf{47.67} &\textbf{ 78.19} & \textbf{49.20 }\\
\bottomrule
\end{tabular}
\vspace{6pt}
\caption{5-shot performance results on five evaluation benchmarks. \textbf{Mi:dm 2.0 Base-8B} refers to the base model before DuS is applied to Mi:dm 2.0 Base-init. \textbf{Mi:dm 2.0 Base-init} is the model after DuS without additional training. \textbf{Mi:dm 2.0 Base*} represents an intermediate checkpoint of Mi:dm 2.0 Base-init, trained on 100B tokens.}
\label{tab4:midm_performance}
\end{table}

\subsubsection{Mi:dm 2.0 Mini: Pruning and Distillation}
\label{section3-1-2}
\begin{figure}[htbp]
  \centering
  \includegraphics[width=0.5\linewidth]{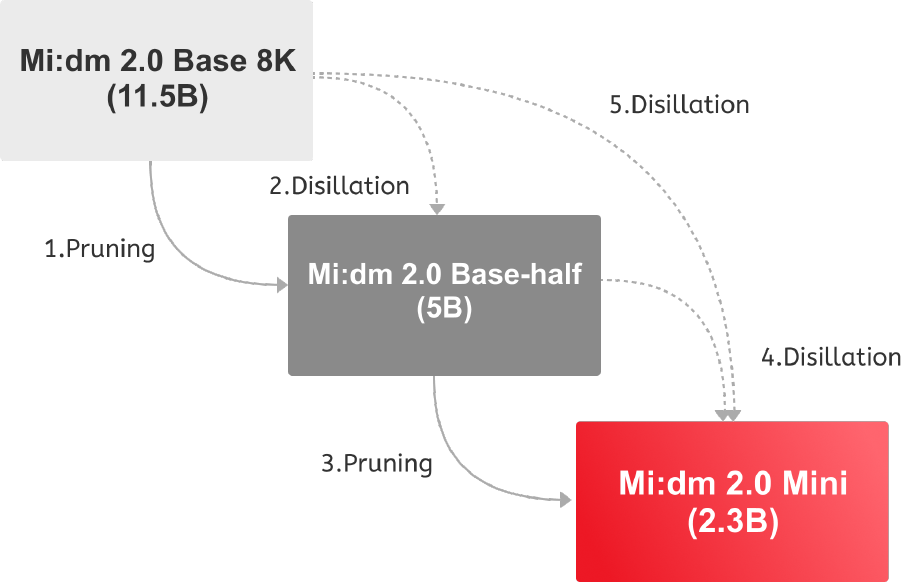}  
  \caption{Pruning and Distillation Workflow of Mi:dm 2.0 Mini}
  \label{fig6:figure7}
\end{figure}

Mi:dm 2.0 Mini is a lightweight variant derived from Mi:dm 2.0 Base, designed to run on low-resource environments such as on-device deployments or low-spec GPUs. To retain the knowledge acquired by Mi:dm 2.0 Base while reducing the model size, Mi:dm 2.0 Mini undergoes two stages of pruning and distillation, as illustrated in~\cref{fig6:figure7}.
In the first stage, we apply width pruning to Mi:dm 2.0 Base to produce an intermediate model of approximately 5B parameters, which we refer to as Mi:dm 2.0 Base-half. This intermediate model then undergoes knowledge distillation, with Mi:dm 2.0 Base acting as the teacher model, guiding the student model to mimicking its output.
In the second stage, we apply further width pruning to this intermediate model. We also adopt a weight-sharing structure for the word embedding to finalize Mi:dm 2.0 Mini's architecture. During this second distillation stage, we use both the intermediate model (Mi:dm 2.0 Base-half) and the Mi:dm 2.0 Base as teacher models, sequentially training the student model to mimic their outputs.

This multi-stage knowledge distillation approach addresses challenges that arise when there is a large capacity gap between Mi:dm 2.0 Base and Mi:dm 2.0 Mini~\cite{muralidharan2024compact, wang2020minilm}. When the disparity between the teacher and the student model size is significant, single-stage distillation often fails to effectively transfer knowledge and can destabilize the training process. Additionally, repeatedly relying on a large teacher model increases the overall computational cost\cite{sreenivas2024llm, muralidharan2024compact}.

To mitigate these challenges and strike a balance between performance and efficiency, we introduce an assistant teacher model—approximately half the size of Mi:dm 2.0 Base—as an intermediary in our pruning-distillation pipeline~\cite{sreenivas2024llm, mirzadeh2020improved}. This intermediate-sized model serves as a critical mediator, enabling more stable and effective knowledge transfer while optimizing both learning performance and computational efficiency.

Furthermore, our width pruning strategy preserves the depth introduced by ~\cite{kim2023solar} during Mi:dm 2.0 Base construction while effectively reducing the overall parameter count. In the subsequent knowledge distillation process, the student model’s initialization and structural design are critical for effectively absorbing knowledge from the teacher. Directly pruning the teacher model for student initialization can cause over-reliance on the original architecture and limit generalization. Prior studies have shown that, for equivalent parameter budgets, deeper models typically outperform wider ones~\cite{sreenivas2024llm, liu2024mobilellm}. Guided by this insight, we adopt a strategy to reduce the width rather than depth when defining the intermediate and final model architectures.

To validate the effectiveness of our pruning-distillation strategy for Mi:dm 2.0 Mini, we conduct proxy model experiments following methodologies in~\cite{bak2025kanana, pengyarn, muralidharan2024compact}. Specifically, we use a 1.8B-parameter model sharing the same architecture as Mi:dm 2.0 Base as a proxy. We compare a scratch-trained version (1.8B-scratch) with a pruned-and-distilled version (1.8B-distill) across major benchmarks, as shown in~\cref{tab5:distill_proxy_benchmark}. Results demonstrate that the distilled model achieves superior performance on most benchmarks while reducing computational cost by approximately 4.5 times lower compared to the scratch-trained baseline.

In both pruning stages, width pruning primarily targets the model’s hidden dimensions and MLP dimensions. For post-pruning calibration, we sample 1,024 examples from Stage-2 pre-training data of Mi:dm 2.0 Base. The same sampling approach is applied to construct the knowledge distillation dataset. Distillation training uses a peak learning rate of 1e-4 with warm-up and cosine decay scheduling, following details described in~\cite{muralidharan2024compact}.

\begin{table}[ht]
\centering
\renewcommand{\arraystretch}{1.1}
\footnotesize
\begin{tabular}{l|cccc}
\toprule
\textbf{Type} & \textbf{MMLU} & \textbf{AGIEval} & \textbf{Winogrande} & \textbf{NQ} \\
 & (acc, 5-shots) & (acc, 5-shots) & (acc, 5-shots) & (EM, 64-shots) \\
\midrule
1.8B-scratch & 26.92 & 18.59 & 64.25 & 9.89 \\
1.8B-distill & 42.49 & 19.55 & 65.27 & 9.36 \\
\midrule\midrule
\textbf{Type} & \textbf{TriviaQA} & \textbf{KMMLU} & \textbf{HAERAE} & \textbf{GSM8K} \\
 & (EM, 64-shots) & (EM, 5-shots) & (acc-norm, 3-shots) & (EM, 5-shots) \\
\midrule
1.8B-scratch & 33.26 & 29.79 & 23.01 & 12.13 \\
1.8B-distill & 34.59 & 32.35 & 52.80 & 32.22 \\
\bottomrule
\end{tabular}
\vspace{6pt}
\caption{Validation results of distillation-based lightweighting using proxy models. The 1.8B-scratch model is trained from scratch, whereas the 1.8B-distill model is distilled from a larger model. \textit{acc} and \textit{EM} denote Accuracy and Exact Match, respectively.}
\label{tab5:distill_proxy_benchmark}
\end{table}

\subsubsection{Long-context Extension}
\label{section3-1-3}
To enable Mi:dm 2.0 to handle long input sequences, we introduce an additional long-context training phase at the final stage of pre-training. This phase extends the model’s maximum input context length from 8,192 tokens to 32,768 tokens.

Following insights from~\cite{men2024base} on the relationship between base frequency in positional encoding and context length, we adjust Mi:dm 2.0’s base frequency from 10,000 to 8 million. Additionally, as highlighted in~\cite{gao2024train}, training with even longer sequences improves performance when targeting context lengths of 32K tokens. Therefore, long-context training uses data with input lengths of up to approximately 65,000 tokens.

Before this final training phase, we conduct experiments to validate data mixing strategies for long-context learning and to mitigate catastrophic forgetting as input length increases. Based on these results, the training dataset is primarily composed of sequences up to 65,000 tokens, with a small proportion of shorter data packed to match this length. Long context training is performed for 2,000 steps using a fixed learning rate of 1e-5 without a separate warm-up phase.

\subsection{Training Costs}
\label{section3.2}
Mi:dm 2.0’s large-scale pre-training is conducted on a high-performance GPU cluster built with Microsoft Azure CycleCloud~\cite{azure_cyclecloud}. This infrastructure is optimized for large-scale computations required in language model training, providing efficient resource management and flexible scalability to support long-term pre-training.

The computing cost at each training stage for each model is summarized in~\cref{tab6:midm_training_flops}. All metrics are reported based on a cluster of NVIDIA H100 GPUs. Floating-point operations (FLOPs) are calculated based on the number of model parameters, sequence length, training steps, and batch size.

\begin{table}[ht]
\centering
\footnotesize
\begin{tabular}{l|c}
\toprule
\textbf{Model Type (Size)} & \textbf{Total Amount of Computation (FLOPs)} \\
\midrule
Mi:dm 2.0 Mini (2.3B) & $4.57 \times 10^{21}$ \\
Mi:dm 2.0 Base (11.5B) & $1.74 \times 10^{23}$ \\
\bottomrule
\end{tabular}
\vspace{6pt}
\caption{Training costs for Mi:dm 2.0. For Mi:dm 2.0 Mini, FLOPs are calculated based only on the student model, excluding the cost of the teacher model.}
\label{tab6:midm_training_flops}
\end{table}

\section{Post-training}
\subsection{Overview}
\label{subsec:Overview}

Pre-trained LLMs possess a wide range of linguistic understanding and generation capabilities based on vast text corpora. However, to achieve the level of reliability required for real-world applications—such as precise instruction-following, logical reasoning, utilization of up-to-date information, tool use, safety, and long-context handling—further fine-tuning is essential. 

Accordingly, the post-training process of Mi:dm 2.0 is designed to systematically enhance six key capabilities critical for maximizing utility and trustworthiness in actual service environments:

\begin{enumerate}[label=\arabic*)]
    \item \textbf{Instruction-Following (IF)}: IF refers to the ability to interpret an instruction accurately and generate responses that match the requested content type, format, length, and structure. Since user queries and demands vary widely in real-world scenarios, strict adherence to instructions is necessary to ensure appropriate information delivery and service quality.

    \item \textbf{Reasoning}: Reasoning refers to the ability to solve complex problems through logical and mathematical thinking, including multi-step operations. These abilities are essential for practical applications and determine the model’s utility in real-world tasks.

    \item \textbf{Retrieval-Augmented Generation (RAG)}: RAG refers to the capability to retrieve external knowledge, documents, or databases in real time and generate responses based on accurate and up-to-date information. This minimizes hallucinations and supports trustworthy decision-making in professional settings.

    \item \textbf{Agent Ability}: Agent ability refers to the capacity to call various tools or APIs via designated interfaces to perform real-world tasks. Beyond simple Q\&A, this is a core competency for service-oriented AI that handles complex scenarios such as scheduling or code execution.

    \item \textbf{Safety}: Safety refers to the ability to ensure social and ethical responsibility, including harmlessness, bias mitigation, and privacy protection. Strengthening this area is vital to safeguarding users and organizations from harmful content, misinformation, or data leakage during deployment.

    \item \textbf{Long Context Handling}: Long context handling refers to the ability to retain and consistently reference important information across long documents, conversations, or code. This ensures coherent and accurate responses in complex tasks such as summarization of lengthy materials.
\end{enumerate}

\begin{figure}[htbp]
  \centering
  \includegraphics[width=0.95\linewidth]{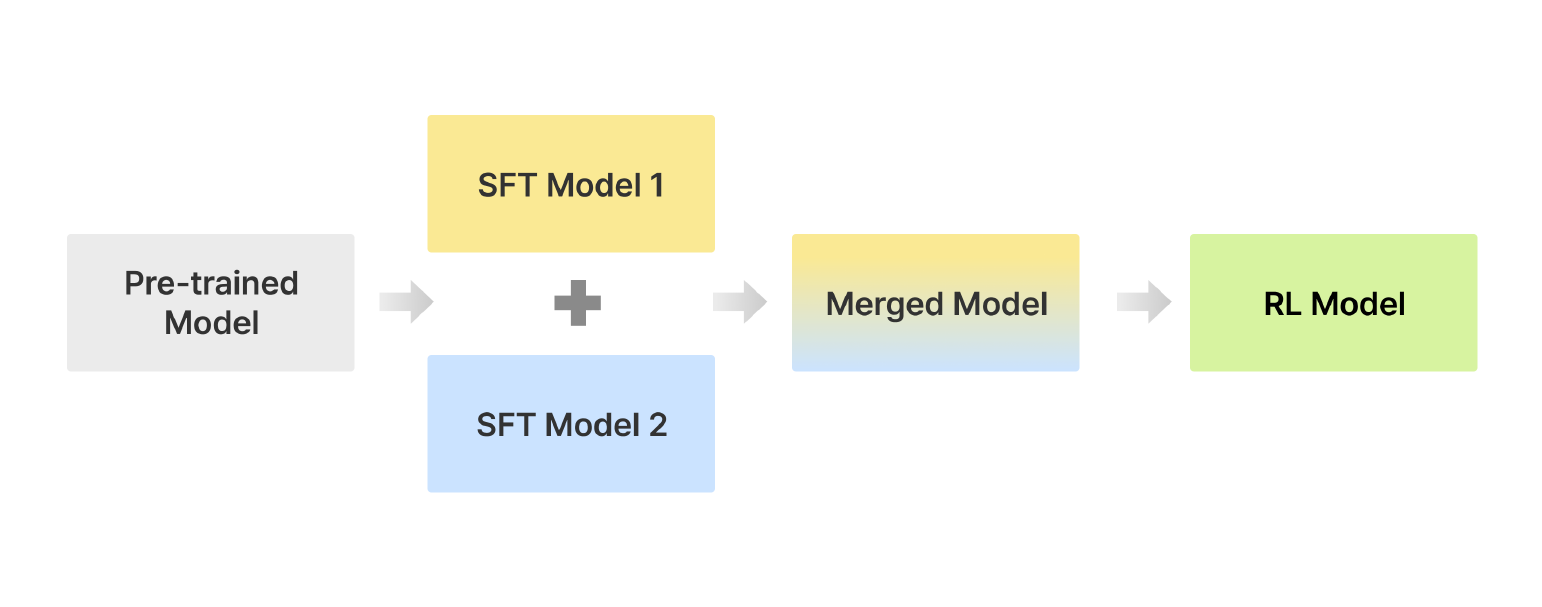}  
  \caption{Post-training process of Mi:dm 2.0}
  \label{fig7:sftmidm-architecture}
\end{figure}

To embed these capabilities, Mi:dm 2.0's post-training consists of structured alignment and specialization strategy to meet real-world demands. The overall post-training process is illustrated in ~\cref{fig:midm-architecture} and is consisted of the following steps: supervised fine-tuning (SFT) on specific tasks, weight merging of multiple SFT models, and final preference optimization through reinforcement learning (RL).

\begin{itemize}
    \item \textbf{SFT}: This step focuses on training the model to to balance generality and specificity, enabling broad responsiveness to user queries.

    \item \textbf{Weight Merging}: This step integrates strengths and features of multiple SFT models via weight merging~\cite{goddard2024arcee}, allowing diverse capabilities acquired through different data and training strategies to be unified in a single model.

    \item \textbf{RL}: This step enhances the model via online/offline reinforcement learning based on human or AI preferences, allowing the model to generate responses that better align with desirable qualities in both content and form.
\end{itemize}

This multi-stage strategy is designed to create synergies across four key aspects: (1) balancing specificity and generality, (2) efficiently integrating individual strengths, (3) improving training efficiency, and (4) aligning with core preferences.

Furthermore, unlike pre-training, post-training must incorporate data structures that meet real service environment requirements. While pre-training typically uses unstructured or single-turn corpora, real conversational services demand multi-turn interactions and role-based dialogue.

Therefore, Mi:dm 2.0 adopts the same chat template format used in LLaMA 4~\cite{meta2025llama} during post-training to reflect multi-turn interaction and role separation. Each utterance is clearly marked by roles such as \texttt{System}, \texttt{User}, \texttt{Assistant}, or \texttt{Tool}. Data is stored as structured multi-turn dialogues rather than simple corpora, and this structure is maintained throughout training and inference.

This structural approach encourages the model to learn realistic conversational scenarios, understand user intent, follow system instructions, and call appropriate tools for each contextual situations. It also improves the model’s ability to maintain consistency, track information, and reference context according to each role.

\subsection{Training Strategy}

\subsubsection{Supervised Fine-Tuning}

SFT is the starting point of Mi:dm 2.0's post-training and plays a key role in equipping the model with real-world capabilities. From a mixture of datasets with diverse objectives and characteristics, we allocate fixed proportions to specific capabilities. Using this curated dataset, multiple SFT models are then trained using supervised learning. After training, their weights are merged so that the diverse capabilities acquired by each model are integrated into a single model. This allows the strengths of each model to be combined effectively.

\begin{figure}[ht]
    \centering
    \includegraphics[width=0.8\textwidth]{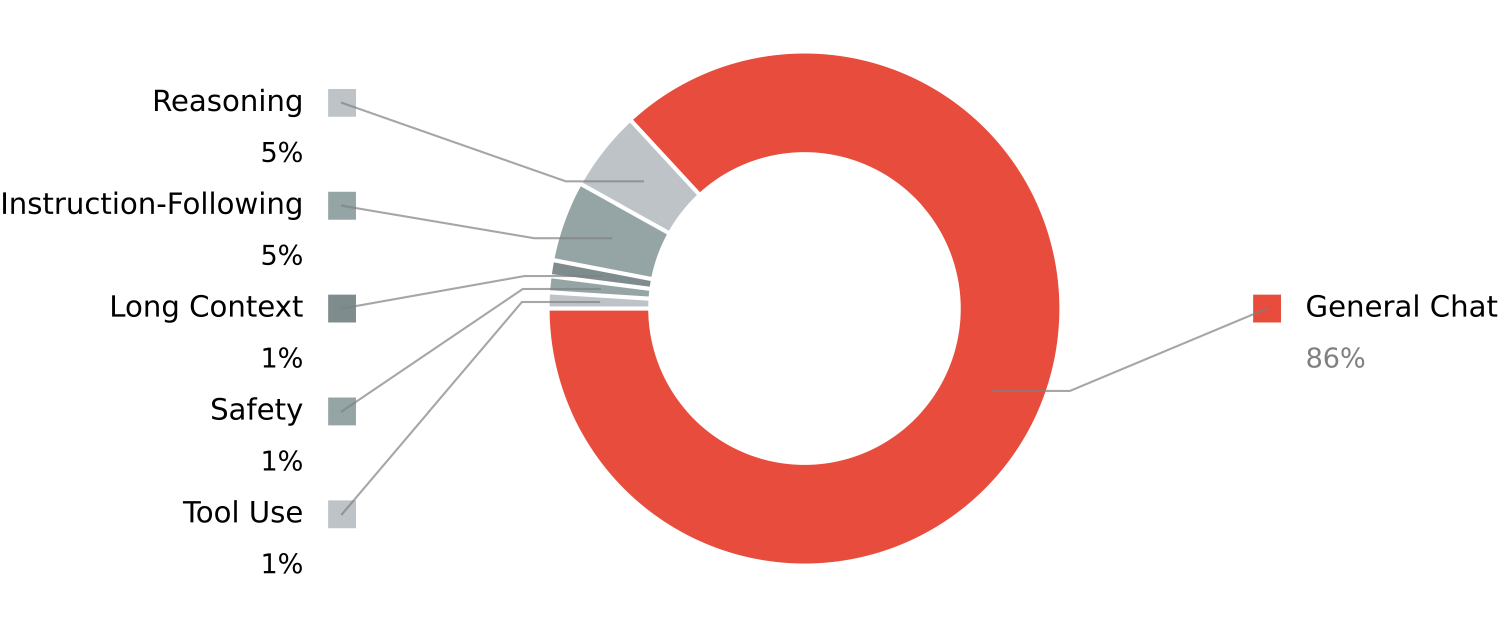}
    \caption{SFT dataset composition ratio (Korean)}
    \label{fig8:sft_data_ko}
\end{figure}

\begin{figure}[ht]
    \centering
    \includegraphics[width=0.8\textwidth]{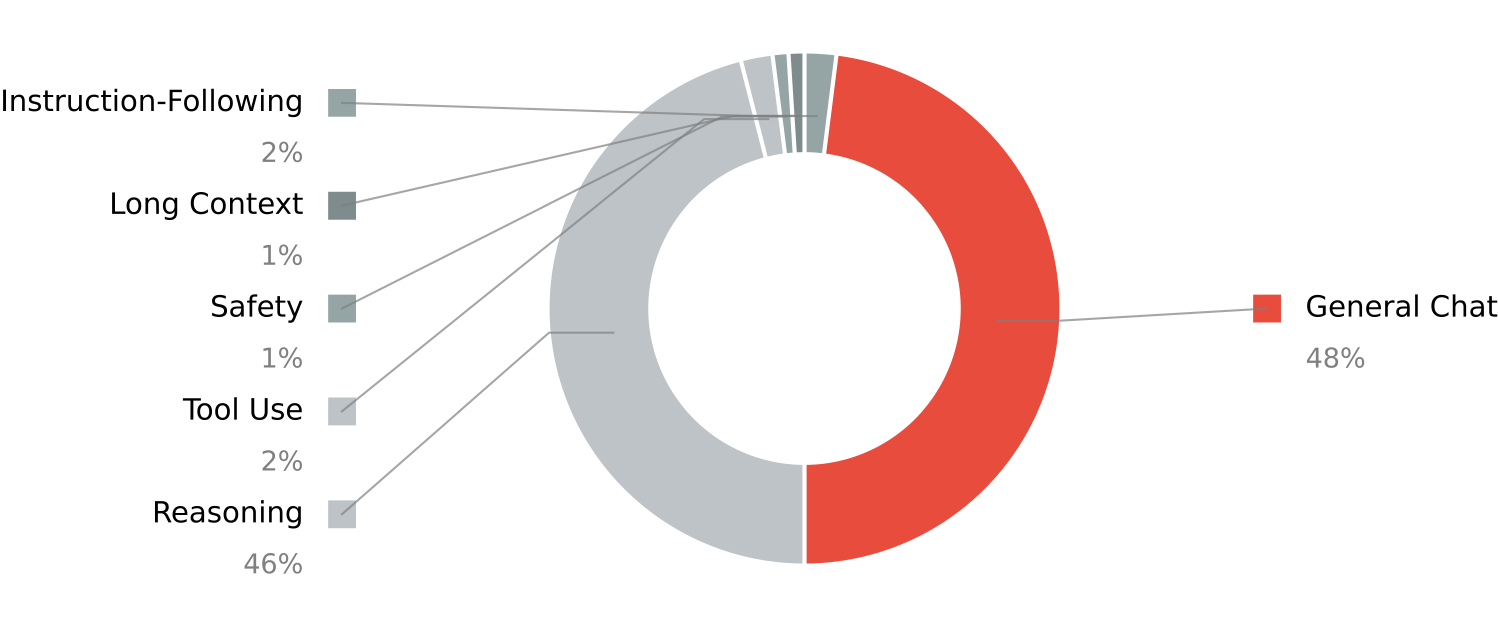}
    \caption{SFT dataset composition ratio (English)}
    \label{fig9:sft_data_en}
\end{figure}

To ensure that the model acquires a well-balanced set of abilities, each training batch is constructed by mixing samples across different capabilities. Definitions of these capabilities and data construction methods are described in \ref{subsec:Capabilities}. To encourage cross-lingual transfer from English (a high-resource language) to Korean (a low-resource language), a portion of translated data is also included in this phase. This batch design aims not only to specialize the model in specific areas but also to broaden its ability to handle a wide range of tasks and complex user demands in real-world service environments. In particular, the complementary nature of English and Korean is fully leveraged, focusing on using high-quality English data to improve Korean performance in a balanced manner. The proportions of Korean and English datasets used in Mi:dm 2.0 SFT training are shown in ~\cref{fig8:sft_data_ko} and ~\cref{fig9:sft_data_en}.

During the SFT phase, the final model is obtained using the training configuration that yields both the most stable and highest-performing results across various experiments. Key considerations include GPU utilization, memory efficiency, and training speed in large-scale multi-node environments. Accordingly, hyperparameters were repeatedly tuned to derive optimal values. The training pipeline incorporates techniques such as data/tensor/pipeline parallelism, which are well-suited for large language model training. The settings for final model merging were also selected based on experimental results comparing multiple weight merge ratios.

\subsubsection{Preference Optimization}

Preference optimization focuses not just on generating correct answers, but also on producing responses that align with actual user experience and expectations. While SFT enables the model to learn patterns and explicit answers from data, real-world applications require the model to generate responses that reflect user preferences, expectations, and intent.

To this end, Mi:dm 2.0 incorporates preference-based training using datasets labeled with human or AI preferences. This approach allows the model to prioritize generating responses that align with key attributes such as safety, reliability, and usefulness. As a result, the model can produce more refined and appropriate responses compared to the baseline SFT model, significantly enhancing its usability, trustworthiness, and user satisfaction.

\begin{figure}[ht]
    \centering
    \includegraphics[width=0.8
    \textwidth]{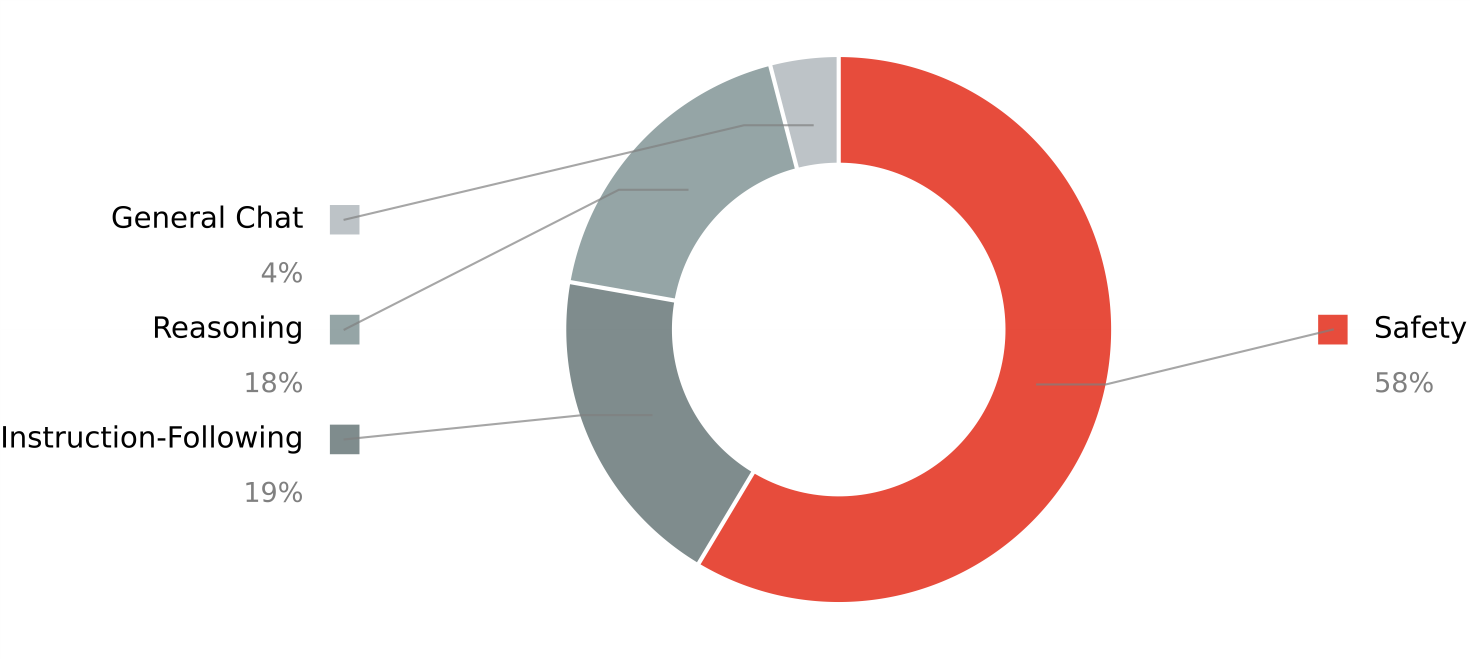}
    \caption{Dataset composition for preference optimization in Mi:dm 2.0}
    \label{fig10:pref_data}
\end{figure}

~\cref{fig10:pref_data} shows the dataset composition used in Mi:dm 2.0’s preference optimization stage. Like the SFT stage, this training phase also employs the state-of-the-art parallelization techniques also conducted in the same large-scale training environment.

\subsection{Capabilities}
\label{subsec:Capabilities}

In this section, we describe the dataset construction strategies developed to effectively embed the core capabilities mentioned in \ref{subsec:Overview}—including IF, reasoning, RAG, agent ability, safety, and long context handling—in addition to general conversation skills. For each capability, we outline the design and implementation of the data generation pipeline, along with the strategies applied in data collection, augmentation, and filtering.

Before exploring each capability in detail, it is important to highlight that, in line with the principle of building a responsible AI, Mi:dm 2.0 excludes datasets with non-commercial licenses from the post-training phase. The post-training dataset for Mi:dm 2.0 consists of a mix of proprietary datasets, alliance-contributed data, and various open-source datasets. Hence, we conducted a thorough review to ensure that any datasets with commercial restrictions or legal uncertainties were excluded. Furthermore, from the perspectives of data quality and ethics, we minimized any legal or societal risks that could arise in the future commercialization of services.

\subsubsection{General Chat}

General Chat refers to the foundational abilities of an AI assistant. It includes a broad range of task capabilities needed in real-world deployment environments, such as open-domain question answering, closed-domain question answering, summarization, writing, transformation, and classification.

Given the wide scope of this domain, we establish a hierarchically structured taxonomy of core skills that Mi:dm 2.0 aims to achieve. This taxonomy consists of 9 major categories (commonsense, knowledge, comprehension, generation, reasoning, instruction-following, multi-turn conversation, multi-step reasoning, long context handling) and 26 subcategories.

To build the training data for general chat functions, we devised a \textit{Core Set Construction Pipeline}, which orthogonally combines the functional subcategories with the domain classification system introduced in \ref{subsection2-2:data} (6 major domains and 20 mid-level domains). Specifically, we created 260 combinations by pairing 20 mid-level domains with 13 functional subcategories for post-training dataset construction. To classify the domain and functional scope of each data instance, we employ filtering and selection strategies based on the embedding similarity.

The filtered dataset within this pipeline is referred to as the \textit{Core Set}. Constructing training data based on this Core Set, as opposed to simply grouping by data source, allows us to include a wide range of functional coverage. This approach ensures that the differences in quality, difficulty, domain, and functional roles of each instance are fully reflected. Consequently, it prevents overrepresentation of specific domains or functions, which could otherwise lead to repeated learning of narrow tasks and induce bias in the model's performance. In practice, using the data constructed with the Core Set pipeline significantly contributes to both training efficiency and robust generalization performance of Mi:dm 2.0.

The major steps of the Core Set Construction Pipeline are as follows:

\begin{enumerate}[label=\arabic*)]
    \item \textbf{Candidate data pre-processing and metadata assignment:} This step involves tagging each data instance with domain and skill metadata, and generating embedding vectors.
    
    \item \textbf{Sample selection based on coverage and diversity:} In this step,  under-represented cells in the domain-skill matrix are filled first. Data with embedding similarity above a certain threshold are considered duplicates and are excluded. Once a predefined number of instances per cell is reached, additional data are no longer added to that cell.
\end{enumerate}

\subsubsection{Instruction-Following}

Instruction-following capability plays a crucial role in enabling the model to accurately understand and reflect diverse user instructions, styles, formats, and constraints. In real service settings, users often present detailed requests beyond simple information delivery, including sentence style, output format, tone, length, and manner. Accurately responding to such demands is an essential requirement for high-performance AI assistants, and IF forms the foundation of this ability.

To enhance IF capability, we first adopt an instruction taxonomy based on IFEval~\cite{zhou2023ifeval}. The instruction types are categorized using various criteria such as style/tone/manner (e.g., honorific, informal, dialectal, humorous), target/situation/role assignment (e.g., age, profession, persona), response language (Korean/English), format and structure (e.g., Markdown, JSON, tables, deductive writing, sentence/paragraph count), inclusion of examples, and constraints (e.g., required/excluded keywords, frequency, length). Based on these, we systemized 37 categories of instruction types.

After establishing the instruction taxonomy, we shift our focus to high-quality, large-scale automatic generation and verification of IF data incorporating realistic user constraints. The IF data pipeline is as follows:

\begin{enumerate}[label=\arabic*)]
    \item \textbf{Initial Query Design with Multiple Constraints}: Initial queries are collected based on frequently encountered real-world tasks (e.g., summarization, transformation, classification, creation, Q\&A). Each query is paired with one or more constraints such as tone, format, role, response length, and language. The total number of constraint combinations reaches 47, representing realistic user instruction patterns. For example: "Summarize the following sentence in table format under 100 characters. Respond in honorific Korean." or "Explain to an elementary school student in English, within 3 sentences."
    \item \textbf{User Request Generation}: Based on the initial queries and constraint combinations, user requests are generated using high-performance LLMs or crowdworkers.
    \item \textbf{Model Response Generation}: All generated user requests are answered using a high-performance LLM.
    \item \textbf{Response Evaluation and Filtering}: Each response is evaluated on two criteria: (i) Satisfaction of constraints (True/False, using a detailed checklist per constraint), and (ii) Response accuracy, completeness, consistency, and appropriateness on a [0.0, 1.0] scale. Responses failing to meet the constraints are excluded from the final dataset.
\end{enumerate}

Additionally, the RL-style IF dataset construction pipeline proceeds as follows:

\begin{enumerate}[label=\arabic*)]
    \item \textbf{Request Generation with Constraints}: Initial queries are augmented with constraints (e.g., "3 sentences or less, table format, honorific tone") to create a large user request set using LLMs and crowdworkers.
    \item \textbf{Logical Consistency Verification}: Each request is checked via LLMs-based scripts for logical contradictions, redundancy, or unrealistic phrasing. Inadequate questions are discarded.
    \item \textbf{Response Pair Construction}: Each question is paired with a chosen/rejected response set. The chosen response serves as the ground truth, while the rejected response is sourced from Mi:dm 1.0 or from generic replies previously attached to the query.
    \item \textbf{Final Question-Answering and Filtering}: A multi-layer quality check is conducted using LLMs-based automatic scoring, heuristic rules, and manual sampling to ensure clarity in preference pairs.
\end{enumerate}

When training the model using this dataset, we observed a 12\%p improvement over the baseline in our own evaluation benchmark KoIFEval, which excludes this dataset.

\subsubsection{Reasoning}

Reasoning capability is essential for solving complex problems, performing mathematical and logical thinking, and executing multi-step tasks. It is one of the core abilities determining the real-world utility of language models. In Mi:dm 2.0’s post-training, math word problems (MWP) at high-school level and code-based problems that combine mathematical and logical reasoning (e.g., MathCoder~\cite{wang2023seamless}-style tasks) were primary learning targets.

In general, the ability to consistently derive correct explanations and final answers in complex math problems, and to logically code, is a key measure of the model’s reliability in service and benchmarking scenarios.

To embed this capability, we construct reasoning datasets. For SFT-style data, we referenced the OpenThoughts~\cite{guha2025openthoughts} approach. First, reasoning-specialized models generate solutions for seed data. Then, the appropriateness and completeness of each solution are verified by using LLMs as judges. After verification, only the final answer portion (excluding the reasoning trace) is extracted and used for SFT training. This method significantly improves model performance in math and reasoning benchmarks. By adding reasoning data to SFT training, performance improved by 7\%p on HRM8K, 13\%p on MATH, and 19\%p on MMMLU compared to the baseline.

In the future, we plan to extend dataset construction to cover scientific and general-domain reasoning beyond mathematics and coding.

\subsubsection{Retrieval-Augmented Generation}

RAG is a core capability that enables a model to retrieve and utilize information from external documents, databases, or real-time knowledge resources to produce more accurate and reliable responses. In Mi:dm 2.0’s post-training, RAG data spans a broad range of general domain documents as well as those tailored for mathematical reasoning. We establish a carefully designed data construction strategy to embed RAG capabilities effectively.

For general document-based RAG datasets, questions are broadly classified into two categories: \textit{factual} and \textit{reasoning}. This classification is reflected in the prompt design. Factual questions refer to those where a single, clearly defined answer exists within the document. In contrast, reasoning questions require synthesizing multiple pieces of information from the document, or involve logical, conditional, or multi-hop reasoning. The construction pipeline is as follows:

\begin{enumerate}[label=\arabic*)]
  \item \textbf{Document Acquisition and Format Conversion}: Documents from various domains (e.g., administration, finance) are collected and converted from PDF to HTML, then to Markdown format through a sequential cleansing process.
  \item \textbf{Keyword Extraction and Question Generation}: Around 10 representative keywords are extracted per document. Based on these keywords, a diverse range of factual and reasoning questions is generated. Prompts used for synthetic data generation include difficulty control (easy, medium, hard) and question-type variation to ensure question diversity.
  \item \textbf{Answer and Evidence Generation}: For each question, file search engine and LLMs are used to generate the answers. These are based either on referenced gold passages or top-N search results, with direct citation of source passages or reasoning traces built from retrieved content.
  \item \textbf{Post-processing and Refinement}: Data containing keywords such as “error” or “recalculate” is removed. Duplicate questions, weak evidence, or computational mistakes are filtered using a multi-step quality check pipeline.
\end{enumerate}

For math-specific RAG data, the process is as follows:

\begin{enumerate}[label=\arabic*)]
  \item \textbf{Document Acquisition and Format Conversion}: Documents such as budgets and statistical reports from education offices, public institutions, or schools are collected and converted from PDF to HTML and then to Markdown.
  \item \textbf{Keyword Extraction and Question Generation}: Math-related keywords (e.g., calculation, statistics) are extracted from each document. Based on these, reasoning questions with adjustable difficulty are created.
  \item \textbf{Answer and Evidence Generation}: Answers are derived either by directly citing numeric, tabular, or formulaic content from the source documents, or by building reasoning traces from search-based results.
  \item \textbf{Post-processing and Refinement}: Low-quality samples (e.g., duplicates, incorrect answers, computational errors) are removed. The final dataset includes a wide range of content such as calculations/statistics, tables/graphs, and rules/scenarios.
\end{enumerate}

This RAG dataset construction pipeline ensures that the model can handle diverse question types across domains while maintaining high data reliability and generalizability through a rigorous quality control standard. By balancing domain-specific and math-focused RAG datasets, the model can respond robustly to real-world complex information queries.

\subsubsection{Tool Use}

Tool use refers to the model’s ability to perform complex user tasks by invoking functions or interacting with external systems. Mi:dm 2.0's tool-use training dataset is built on function call-based dialogues that conform to the model context protocol (MCP) standard. This enables Mi:dm 2.0 to act as an AI agent that can select, call, and manage multiple tools in real-world situations, processing information in multiple steps and meeting user needs effectively. Key capabilities include tool selection, intent understanding, scenario variation, multi-tool coordination, and exception handling (e.g., missing tools, argument errors).

The tool-use dataset includes 249 functions across 40 real-life topics such as health/fitness, food/cooking, finance, workplace, tech devices, and emotion management. Each tool is defined with a JSON schema including its name, description, input/output formats, and example or enumerated values. The design closely mimics real-world API structures. Example tools include calorie calculation, currency conversion, emotion detection, and device search.

The dataset construction process is as follows:

\begin{enumerate}[label=\arabic*)]
  \item \textbf{Tool Definition and Listing}: Tools are designed per topic, each defined with JSON schemas detailing input arguments, output types, and examples.
  \item \textbf{Scenario and Persona Generation}: Diverse user personas are defined (age, gender, occupation), and realistic subtopics and scenarios are created based on these personas and the tools associated with each topic. Each scenario may involve 1–3 tools relevant to the user’s context.
  \item \textbf{Multi-turn Dialogue Generation}: Each scenario leads to multi-turn conversations, consisting of user queries and model responses. Model responses fall into six categories: additional information requests, tool invocation, tool execution results, responses based on tool outputs, casual/chit-chat, and lack of tool support. For instance, if a user asks for a device recommendation, the model might first ask for brand preferences, call the tool, and then summarize the result.
  \item \textbf{Data Cleaning}: Rule-based filters remove entries with undefined functions, incorrect arguments, mismatched response types, unnecessary repetition, typos, or inaccurate answers.
\end{enumerate}

\subsubsection{Safety}

Safety is one of the most important requirements when applying language models in the real world. Mi:dm 2.0 evaluates safety through three core principles: harmlessness, honesty, and consistency in AI role behavior. Harmlessness is categorized into 7 major classes—sexual content, violations, violence, bias/discrimination, politics, disasters, and profanity—further divided into 56 subcategories. Honesty includes avoiding false or misleading information, especially in expert domains like healthcare, law, and finance. Consistency in AI role entails maintaining coherent identity and purpose, refusing to anthropomorphize itself, and avoiding inappropriate role-play.

Mi:dm 2.0 applies these detailed standards to enhance social and ethical alignment, reduce harmful or biased responses, protect personal data, and ensure the appropriateness of refusal or deflection replies in both service and experimental settings. In particular, Mi:dm 2.0 incorporates sophisticated guidelines for borderline queries—those whose harmfulness depends heavily on context or language.

The safety training dataset includes both open-ended question-answering (QA) pairs and closed multiple-choice formats. The open-ended data covers social taboo, ethical, or legally risky topics, and borderline prompts involving ambiguity, metaphor, or contextual shifts. The closed data consists of well-defined multiple-choice questions designed to test honesty, harmfulness detection, and norm compliance.

The data construction pipeline is as follows:
\begin{enumerate}[label=\arabic*)]
  \item \textbf{SFT Data}: Templates for rejections, rejection reasons, expert referrals, guidance statements, and AI role declarations are standardized, categorized, and applied across different prompt types.
  \item \textbf{RL Data}: Each prompt is associated with multiple pairs of chosen and rejected responses, facilitating training for both preference optimization and accurate safety handling.
\end{enumerate}

Incorporating these datasets during both the SFT and RL phases lead to significant improvements: enhanced quality and diversity of refusal responses, reduced over-rejection of borderline queries, improved honesty and consistency, more effective rejection of unsafe responses during deployment, and stronger privacy protections. Notably, models trained with RL-based safety data demonstrated a 24\%p improvement in safety performance compared to those without such training.

\subsubsection{Long Context}

As in pre-training, post-training also emphasizes the ability to handle long context. From a post-training perspective, long context capabilities refer to effectively performing tasks such as information retrieval, comprehension, reasoning, and summarization within lengthy input documents. These abilities are essential in practical applications like document summarization, complex QA, dialogue systems, code analysis, and planning.

In Mi:dm 2.0, a large-scale synthetic dataset in both Korean and English is constructed for long context training. The full process is as follows:

\begin{enumerate}[label=\arabic*)]
  \item \textbf{Corpus Chunking}: Long documents are selected from the pre-training corpus. Documents are split into segments ranging from 4K to 32K tokens at 1K-token intervals. For each token range, 2.2K documents per language are collected.
  \item \textbf{Mini-Context Sampling and Positional Diversity}: From each document, 400-token mini-contexts are extracted at positions corresponding to 0\%, 10\%, 20\%, ..., 100\% of document length. This mitigates positional bias in information access and boosts question coverage in later sections using an exponential sampling strategy.
  \item \textbf{QA Generation and Validation}: LLMs generate questions and answers based on mini-contexts. Only QA pairs scoring 9 or higher (out of 10) by an evaluation model are retained, ensuring high data quality.
\end{enumerate}

Training with this dataset lead to outstanding long-context retrieval and comprehension performance on benchmarks such as Needle-in-a-haystack~\cite{nelson2024needle} and RULER~\cite{hsiehruler}, significantly outperforming models that focused solely on short-context understanding.

\section{Evaluation}
We conduct multi-dimensional evaluations to assess not only global English benchmarks but also the cultural context and real user experience of Korean language use. In bilingual (Korean-English) LLM evaluations, existing global benchmarks have traditionally been biased toward English-centric assessment~\cite{jin2024kobbq,jin2025social,zheng2025rsafe}. To address this, we evaluate Korean-specific capabilities through both quantitative and human evaluation components designed to reflect Korea’s unique linguistic and cultural characteristics. Furthermore, to support Responsible AI, we explicitly evaluate safety and robustness. Through these evaluations, we identify the specific strengths of our model in handling Korean language tasks.


\subsection{Quantitive Evaluation}
We conduct a quantitative evaluation using a diverse set of benchmarks, including publicly available benchmarks, translated English benchmarks, and a proprietary benchmark developed by KT. This comprehensive evaluation specifically assesses the various linguistic and cultural dimensions crucial for Korean language models. To ensure fair comparisons and enhance the reliability and validity of our findings, we applied statistical significance testing. This rigorous approach allows us to confidently verify the inherent performance differences between LLMs.

 \subsubsection{General English Benchmark}
To verify global comparability for Mi:dm 2.0, we include parallel evaluations of its English-language performance. These benchmarks measure not only general language understanding but also performance across various domains, including reasoning, mathematical ability, and specialized knowledge. Evaluation metrics follow the criteria defined by each benchmark, ensuring consistency and alignment with established standards for LLM assessment.

In particular, we select benchmarks for Mi:dm 2.0 evaluation from Hugging Face Leaderboard v2. This choice is motivated by the fact that portions of the evaluation data in Leaderboard v1 had been made public, potentially leading to model overfitting or distorted evaluation results. To mitigate these concerns, we adopt the v2 benchmark as a more robust and reliable evaluation standard. This approach ensures representativeness, validity, and reliability in the evaluation process.

Selected English common ability benchmarks are as follows.
\begin{itemize}
    \item \textbf{Instruction Following} – Evaluates the LLM's ability to execute given commands accurately, measured using the IFEval~\cite{zhou2023ifeval} dataset.

    \item \textbf{Reasoning} – Assesses multi-step logical reasoning across diverse domains using the MuSR~\cite{sprague2024musr}, GPQA~\cite{rein2024gpqa}, and BBH~\cite{bigbench2022bbh} datasets.

    \item \textbf{Mathematics} – Tests problem-solving and step-by-step calculation ability on elementary to high school-level problems using the GSM8K~\cite{cobbe2021gsm8k} dataset.

    \item \textbf{Coding} – Evaluates beginner-level coding skills using MBPP+~\cite{aggarwal2023mbpp+} dataset to assess Python code generation skills.

    \item \textbf{General Knowledge} – Measures understanding and application of specialized knowledge in fields such as science, technology, humanities, and social sciences, using the MMLU~\cite{hendrycks2021mmlu} and MMLU-PRO~\cite{wang2024mmlupro} datasets.
\end{itemize}




\begin{table}[ht]
\centering

\renewcommand{\arraystretch}{1.3}
\resizebox{\linewidth}{!}{\Huge
\begin{tabular}{
l
|c
|cccc
|c|c
|ccc
}
\toprule
\multirow{2}{*}{\textbf{Model}} 
& \makecell{\textbf{Instruction}\\\textbf{Following}} 
& \multicolumn{4}{c|}{\textbf{Reasoning}} 
& \makecell{\textbf{Mathematics}} & \makecell{\textbf{Coding}} 
& \multicolumn{3}{c}{\textbf{General Knowledge}} \\
\cmidrule(lr){2-2} \cmidrule(lr){3-6} \cmidrule(lr){7-7} \cmidrule(lr){8-8} \cmidrule(lr){9-11}
& IFEval 
& BBH & GPQA & MuSR & Avg. 
& GSM8K & MBPP+
& MMLU-pro & MMLU & Avg. \\
\midrule
Qwen3-4B~\cite{yang2025qwen3}
& \underline{79.7} & \textbf{79.0} & \textbf{39.8} & \textbf{58.5 }& \textbf{59.1} & \textbf{90.4} &\textbf{62.4} &  - & \textbf{73.3} &\textbf{ 73.3} \\
Exaone-3.5-2.4B-inst~\cite{exaone3.5_2024} 
& \textbf{81.1} & \underline{46.4} & \underline{28.1} & 49.7 & \underline{41.4} & 82.5 & 59.8 & - & \underline{59.5} & \underline{59.5} \\
\rowcolor{gray!10}
Mi:dm 2.0 Mini-inst 
& 73.6 & 44.5 & 26.6 & \underline{51.7} & 40.9 & \underline{83.1} & \underline{60.9} &- & 56.5 & 56.5 \\ \midrule
Qwen3-14B~\cite{yang2025qwen3} 
& \underline{83.9} & \textbf{83.4} & \textbf{49.8} & \textbf{57.7} & \textbf{63.6} & \underline{88.0} & 73.4& \textbf{70.5} & \textbf{82.7} & \textbf{76.6} \\
Llama-3.1-8B-inst~\cite{meta2025llama} 
& 79.9 & 60.3 & 21.6 & 50.3 & 44.1 & 81.2 & \textbf{81.8}& 47.6 & 70.7 & 59.2 \\
Exaone-3.5-7.8B-inst~\cite{exaone3.5_2024} 
& 83.6 & 50.1 & 33.1 & 51.2 & 44.8 & 81.1 & \underline{79.4}& 40.7 & 69.0 & 54.8 \\
\rowcolor{gray!10}
Mi:dm 2.0 Base-inst 
& \textbf{84.0} & \underline{77.7} & \underline{33.5} & \underline{51.9} & \underline{54.4} & \textbf{91.6} & 77.5& \underline{53.3} & \underline{73.7} & \underline{63.5} \\
\bottomrule
\end{tabular}
}
\vspace{6pt}
\caption{Combined English Benchmark Performance for Mi:dm 2.0 Base and Mi:dm 2.0 Mini Compared to Baseline Models, Including Detailed Subtasks and Category Averages. \textbf{Bold} scores indicate the best performance, and \underline{underlined} scores mean the second best.}
\label{tab7:midm_base_mini_combined}
\end{table}

\cref{tab7:midm_base_mini_combined} summarizes the English benchmark performance of the Mi:dm 2.0 lineup alongside baseline models.
Mi:dm 2.0 Base demonstrates the highest performance in \textit{instruction following} and \textit{mathematics} across all evaluated models, indicating strong capability in these tasks. Moreover, compared to the domestic baseline Exaone-3.5-7.8B-inst, Mi:dm 2.0 Base achieves substantial improvements of 9.6\%p in \textit{reasoning} and 8.7\%p in \textit{general knowledge}. Additionally, Mi:dm 2.0 Mini exhibits comparable performance to the domestic benchmark model (Exaone-3.5-2.4B-inst) in both the \textit{mathematics} and \textit{reasoning} categories.



 \subsubsection{Korean Specific Benchmark}
To evaluate Mi:dm 2.0's understanding of Korean, we develop evaluation metrics specifically designed to capture the language's unique linguistic and cultural features. Existing language model benchmarks are predominantly English-centric and fail to account for essential aspects of Korean, such as honorific forms, Sino-Korean vocabulary, and idiomatic expressions. These gaps limit accurate assessment of Korean language proficiency when relying on English-based evaluation systems.

Furthermore, existing Korean benchmarks are limited in scope and often rely on direct translations from English datasets, introducing distortions and failing to measure authentic language comprehension. To address these limitations and rigorously validate Mi:dm 2.0's performance as a model optimized for Korean, we design dedicated evaluation metrics and construct proprietary high-quality benchmarks at KT.

We design these benchmarks to reflect the structural and semantic complexity of Korean, as well as its broader social and cultural context, enabling more precise and realistic performance assessment. Instead of relying on existing public datasets, we develop original evaluation tasks and domains aligned with real-world Korean usage. Collaboration with domain experts, including researchers specializing in the Korean language and culture, ensures the reliability and domain specificity of these evaluation materials.

The evaluation comprises five main categories: Instruction Following in Korean, Korean Comprehension, Korean Reasoning, Korean Society and Culture, and Korean General Knowledge. This evaluation enables a comprehensive assessment of linguistic competence and the capacity to handle culturally and contextually appropriate content.

Each evaluation metric is based on the following benchmark datasets. 

\begin{itemize}
    \item \textbf{Instruction Following} – Evaluates the model's ability to follow Korean-language instructions accurately. We utilize the Ko-IFEval dataset~\cite{zhou2023instructionfollowingevaluationlargelanguage}, which categorizes instructions by type, allowing for a detailed analysis of responses. Ko-MTBench dataset~\cite{lgai2024exaone} is used to specifically assess the model's performance on more open-ended and complex conversational instructions. This approach moves beyond simple accuracy rates by providing granular diagnostics of strengths and limitations for each instruction type. 

    \item \textbf{Korean Comprehension} – Tests the understanding of Korean-specific linguistic features, including honorific forms, Sino-Korean vocabulary (words of Chinese origin used in Korean), native words, proverbs, and idiomatic expressions. Benchmarks include K-Pragmatics\footnotemark, K-Pragmatics-hard\footnotemark,, KoBest (BoolQ, SentiNeg)~\cite{park2023kobest}, and Ko-Sovereign\footnotemark, (language and literature domains).

    \item \textbf{Korean Reasoning} – Assesses contextual understanding and logical inference capabilities using pairs of semantically similar sentences. This includes tasks designed to capture nuanced contextual reasoning. Datasets include Ko-Winogrande~\cite{kowinogrande2024}, KoBest (COPA, HellaSwag, WiC), Logic Kor~\cite{ma2024korbench}, and HRM8K~\cite{ko2025understand}.

    \item \textbf{Korean Society and Culture} – Measures referential reasoning skills that require awareness of Korean social and cultural contexts. Data sources include K-Referential\footnotemark, K-Referential-hard\footnotemark, Ko-Sovereign\footnotemark (culture, folklore, and society domains), and HAERAE-bench~\cite{son2023haerae}.

    \item \textbf{Korean General Knowledge} – Evaluates expertise in 45 specialized domains spanning Korean humanities and social sciences, natural sciences, law, and economics. This category is supported by KMMLU~\cite{son2025kmmlu} and Ko-Sovereign datasets, covering history, law, economics, politics, education, and geography.
\end{itemize}
\footnotetext{KT proprietary benchmark, internally developed for Korean-specific evaluation}

The evaluation of Mi:dm 2.0's Korean-language capability incorporates three self-developed, Korean-specific benchmark sets: K-Pragmatics\footnotemark, K-Referential\footnotemark, and Ko-Sovereign\footnotemark. 

K-Pragmatics and K-Referential are designed to assess understanding of Korean-specific linguistic features such as honorifics, Sino-Korean vocabulary, native terms, and proverbs, as well as the ability to perform inferences grounded in Korean social and cultural contexts. Each benchmark also includes "Hard" versions with more challenging questions, enabling the evaluation of model performance limits and robustness on complex tasks.

Ko-Sovereign is KT’s proprietary benchmark developed in collaboration with the Research Institute of Korean Studies at Korea University, ensuring academic rigor through expert consultation with Korean studies scholars. It offers a comprehensive evaluation across nine domains—such as language, culture, history, law, and economics—to measure model expertise in a wide range of Korean social and cultural contexts.

\begin{table}[ht]
\centering
\renewcommand{\arraystretch}{1.2}
\resizebox{\linewidth}{!}{\LARGE
\begin{tabular}{
l
|ccccc
|cccccc
}
\toprule
\multicolumn{11}{c}{\textbf{Comprehension and Reasoning}} \\
\midrule
\multirow{2}{*}{\textbf{Model}} 
& \multicolumn{5}{c|}{\textbf{Comprehension}}
& \multicolumn{6}{c}{\textbf{Reasoning}}\\
\cmidrule(lr){2-6}\cmidrule(lr){7-12}
& K-Prag* & \makecell{K-Refer\\-Hard*} & Ko-Best & \makecell{Ko\\-Sovereign*} & Avg.
& \makecell{Ko\\-Winogrande}& Ko-Best & LogicKor & HRM8K & \multicolumn{2}{c}{Avg.}\\
\midrule

Qwen3-4B              & \textbf{73.9} & \underline{56.7} & \textbf{91.5} & \textbf{43.5} & \textbf{66.6} & \textbf{67.5} & \textbf{69.2} & 5.6 & \textbf{56.7} & \multicolumn{2}{c}{\textbf{43.8}} \\
Exaone-3.5-2.4B-inst & 68.7 & \textbf{58.5} & \underline{87.2} & 38.0 & \underline{62.5} & 60.3 & 64.1 & \underline{7.4} & 38.5 & \multicolumn{2}{c}{36.7} \\
\rowcolor{gray!10}
Mi:dm 2.0 Mini-inst   & \underline{69.5} & 55.4 & 80.5 & \underline{42.5} & 61.9 & \underline{61.7} & \underline{64.5} & \textbf{7.7} & \underline{39.9} & \multicolumn{2}{c}{\underline{37.4}} \\
\midrule

Qwen3-14B             & \textbf{86.7} & \textbf{74.0} & \underline{93.9} & \underline{52.0} & \textbf{76.8} & \textbf{77.2} & \textbf{75.4 }& \underline{6.4} & \textbf{64.5} & \multicolumn{2}{c}{\textbf{48.8}} \\
Llama-3.1-8B-inst    & 59.9 & 48.6 & 77.4 & 31.5 & 51.5 & 40.1 & 26.0 & 2.4 & 30.9 & \multicolumn{2}{c}{19.8}  \\
Exaone-3.5-7.8B-inst  & 73.5 & 61.9 & 92.0 & 44.0 & 67.2 & 64.6 & 60.3 & \textbf{8.6} & 49.7 & \multicolumn{2}{c}{39.5} \\
\rowcolor{gray!10}
Mi:dm 2.0 Base-inst   & \underline{86.5} & \underline{70.8} & \textbf{95.2} & \textbf{53.0} & \underline{76.1} & \underline{75.1}& \underline{73.0} & \textbf{8.6} & \underline{52.9} & \multicolumn{2}{c}{\underline{44.8}}  \\

\midrule  \midrule

\multicolumn{11}{c}{\textbf{Society \& Culture, General Knowledge, and Instruction Following}} \\
\midrule
\multirow{2}{*}{\textbf{Model}} 
& \multicolumn{5}{c|}{\textbf{Society \& Culture}}
& \multicolumn{3}{c|}{\textbf{\makecell{General\\Knowledge}}} & \multicolumn{3}{c}{\textbf{\makecell{Instruction\\Following}}}\\
\cmidrule(lr){2-6}\cmidrule(lr){7-9}\cmidrule(lr){10-12}
& K-Refer* & \makecell{K-Refer\\-Hard*} & \makecell{Ko\\-Sovereign*} & HAERAE & Avg. 
& KMMLU & \makecell{Ko\\-Sovereign*} & Avg. & \multicolumn{1}{|c}{\makecell{~~Ko\\-IFEval~~}} & \makecell{~~Ko\\-MTBench~~}& Avg.\\

\midrule

Qwen3-4B              & 53.6 & 42.9 & 35.8 & 50.6 & 45.7 & \textbf{50.6} & \textbf{42.5} & \textbf{46.5} & \multicolumn{1}{|c}{\textbf{75.9}}  & \underline{63.0} & \underline{69.4}\\
Exaone-3.5-2.4B-inst & \underline{64.0} & \textbf{67.1} & \textbf{44.4} & \underline{61.3} & \textbf{59.2} & 43.5 & \underline{42.4} & 43.0 & \multicolumn{1}{|c}{65.4}& \textbf{74.0}& 68.9\\
\rowcolor{gray!10}
Mi:dm 2.0 Mini-inst   & \textbf{66.4} & \underline{61.4} & \underline{36.7} & \textbf{70.8} & \underline{58.8}& \underline{45.1} & \underline{42.4} & \underline{43.8} & \multicolumn{1}{|c}{\underline{73.3}}  & \textbf{74.0}& \textbf{73.6}\\
\midrule

Qwen3-14B             & \underline{72.4} & 65.7 & \underline{49.8} & 68.4 & 64.1 & \underline{55.4} & \underline{54.7} & \underline{55.1} & \multicolumn{1}{|c}{\textbf{83.6}}  & 71& \underline{77.3}\\
Llama-3.1-8B-inst    & 43.2 & 36.4 & 33.8 & 49.5 & 40.7 & 33.0 & 36.7 & 34.8 & \multicolumn{1}{|c}{60.1}  & 57& 58.5\\
Exaone-3.5-7.8B-inst  & 71.6 & \underline{69.3} & 46.9 & \underline{72.9} & \underline{65.2} & 52.6 & 45.6 & 49.1 & \multicolumn{1}{|c}{69.1}  & \underline{79.6}& 74.4\\
\rowcolor{gray!10}
Mi:dm 2.0 Base-inst   & \textbf{89.6} & \textbf{86.4} & \textbf{56.3} & \textbf{81.5} & \textbf{78.4} & \textbf{57.3} & \textbf{58.0 }& \textbf{57.7 }& \multicolumn{1}{|c}{\underline{82}} & \textbf{89.7}& \textbf{85.9}\\

\bottomrule
\end{tabular}
}
\vspace{4pt}
\caption{Unified Korean Benchmark Results for Mi:dm 2.0 and Comparison Models. * indicates KT proprietary evaluation resources.}
\label{tab7:korean_benchmark_unified}
\end{table}

\cref{tab7:korean_benchmark_unified} shows that the performance of Korean-focused LLMs is comparable to or better than that of global LLMs. Both Mi:dm 2.0 Base and Exaone-3.5-7.8B demonstrate performance equivalent to or exceeding that of similarly sized global models such as Qwen3-14B and Llama-3.1-8B. These results indicate that KT's Korean-specific evaluation metrics and benchmarks are effective in capturing the challenges of the Korean language.

In particular, Mi:dm 2.0 Base achieves the highest overall scores among the comparison models in the Korean society \& culture and general knowledge categories. It outperforms all other models on every benchmark related to Korean society, culture, and specialized knowledge. Notably, on the K-Referential-hard benchmark, it shows a significant performance advantage of 17.1\%p over Exaone-3.5-7.8B, highlighting its superior ability to handle more complex, culturally grounded tasks.

Our model also outperforms all comparison models on KMMLU, demonstrating a particularly significant performance gap of 16\%p in the Korean history subdomain compared to Qwen3-14B, which further validates its profound understanding of Korean history.

Additionally, in the newly developed Ko-Sovereign benchmark, Mi:dm 2.0 Base achieves the highest performance across all nine evaluated domains, demonstrating its comprehensive understanding of Korean-specific knowledge.

Mi:dm 2.0 Mini achieved the highest performance in the instruction-following domain among all compared models. Notably, on the Ko-MTBench, it demonstrated a significant performance advantage of 11\%p over Qwen3-4B, despite posessing only half the number of parameters.

Mi:dm 2.0 Mini also shows strong results in the Korean society and culture category. Specifically, on the HAERAE benchmark, it exceeds the performance of the global comparison model Qwen3-4B by over 20\%p , indicating robust capability for understanding complex social and cultural contexts in Korean.

\subsection{Human Evaluation}
We perform targeted human evaluations on Mi:dm 2.0 to assess dimensions that quantitative metrics cannot fully capture. These qualitative assessments focus on linguistic fidelity and contextual appropriateness in tasks such as Korean-English translation accuracy, the preservation of meaning in summarization, and alignment with expected topics and formats in writing. Rather than measuring simple correctness, the evaluation emphasizes response quality and user-aligned satisfaction to ensure the model meets practical deployment requirements.

We divide the human evaluation into five top-level categories: generation, conversation/question-answering (QA), understanding, analysis/classification, and mathematics. Each category includes 15 total sub-areas. For each sub-area, we define three to five targeted evaluation metrics to assess response quality from multiple dimensions. Example metrics include information accuracy, instruction adherence, and language proficiency, all designed to match the specific goals of each evaluation area.

We evaluate in a blind setup. Three independent evaluators score each question without knowing which model produced the response, using a standardized scale for each metric. We only include results with consistent agreement among evaluators in the final analysis. The weighted aggregation of these scores provides clear and comparable model-level performance metrics.

Qualitative evaluation often produces results that differ from quantitative benchmarks. For example, even if one model achieves a higher quantitative score, human reviewers may identify issues such as mixed-language output or incoherent sentences that lower its qualitative rating. These differences highlight why both evaluation methods are necessary and complementary, ensuring a more complete assessment of model performance.

However, qualitative evaluation has limitations in terms of evaluator subjectivity and the difficulty of ensuring consistency among evaluators. In Mi:dm 2.0 evaluation process, to overcome these limitations, multiple evaluators are involved to ensure consistency among evaluators, and statistical tests are conducted on evaluation results to verify reliability. Additionally, evaluator training, standardization of evaluation processes, and management of the evaluation system are implemented to maintain consistency and systematicity in evaluation procedures. 

The Mi:dm 2.0 Base demonstrated a clear performance advantage in the OpenQA task~\cite{chen2017reading}, outperforming the Qwen3-14B model by 15.2\%p, with scores of 89.5 and 74.3, respectively. This notable difference is not merely a quantitative superiority, but rather a reflection of Mi:dm 2.0 Base's inherent capabilities.

The OpenQA task is specifically designed to evaluate a model's ability to answer a wide array of questions spanning diverse domains, including society, culture, economy, law, history, and language. Crucially, it demands a profound understanding of Korean contexts and topics, along with the capacity to generate accurate and comprehensive responses. Success in this task goes beyond simple factual recall; it necessitates the model's ability to interpret question intent, synthesize information from various sources, and construct logically coherent answers.

The exceptional performance of Mi:dm 2.0 Base in the OpenQA task underscores its deep language comprehension and reasoning abilities, which are cultivated through extensive training on vast Korean datasets. This indicates our model's proficiency in accurately grasping complex relationships, subtle nuances, and cultural specificities embedded within Korean text, and subsequently generating novel information based on this understanding. These results unequivocally establish Mi:dm 2.0 Base's distinctive competence in solving problems based on multi-dimensional Korean knowledge. Consequently, it positions Mi:dm 2.0 Base as a highly reliable language model capable of effectively addressing complex queries within the Korean linguistic environment.

\subsection{RAI Evaluation}
We conduct thorough Responsible AI (RAI) evaluation to ensure and verify Mi:dm’s safety and reliability in real-world use. This evaluation addresses both safety and robustness dimensions. Safety evaluation follows KT-defined AI risk categories and includes structured procedures to identify risks such as harmful content generation (Content Safety Risks), misuse in social and economic contexts (Socio-Economic Risks), and potential rights violations or legal issues (Legal and Rights-Related Risks). The approach combines scenario-based assessments with benchmark-driven evaluations. Robustness evaluation focuses on testing how well the model withstands various adversarial attack techniques that malicious users might try, using dedicated red-teaming methods.

\subsubsection{Scenario-Based Evaluation} 
To qualitatively evaluate model behavior against AI risks, we define detailed topics and keywords for each risk category and design prompts that reflect diverse, realistic user scenarios when interacting with AI services. For example, we evaluate how the model handles requests for methods to collect personal information or demands to justify harmful values. We label responses as safe or unsafe, assign severity scores to harmful outputs, and apply clear evaluation criteria for each risk category to ensure consistency. In addition to harmfulness detection, we also review models for excessive refusal patterns to improve the overall safety evaluation. For this process, we use an internally developed Korean dataset based on XSTest~\cite{rottger-etal-2024-xstest}.


We apply several methods to ensure the reliability of qualitative evaluations. First, multiple evaluators independently review the same responses using a cross-validation approach, and we measure agreement with Fleiss’ kappa coefficient, a standard metric for assessing inter-rater reliability. Second, we strengthen the evaluation process with a secondary verification stage that uses Judge LLMs. By comparing the judgments of two different Judge LLMs with human evaluations based on risk-specific criteria, we verify the consistency of results. Third, we maintain clear and stable evaluation standards through regular training and review sessions for evaluators. As risk criteria evolve and AI responses become more complex, these sessions also help refine and expand evaluation guidelines.


The evaluation uses two metrics: Not Unsafe Rate (\%) and Not Over-Refuse Rate (\%). These measures are the proportion of safe responses and the proportion that avoid excessive refusal, respectively, across all evaluation prompts. Rather than averaging scores by risk category, the overall score is calculated based on the total count of responses that are either safe or do not exhibit excessive refusal across all evaluation items.

\begin{table}[ht]
\centering
\renewcommand{\arraystretch}{1.2}
\resizebox{0.8\linewidth}{!}{\footnotesize
\begin{tabular}{lcccc}
\toprule
\textbf{Model} & \textbf{Content Safety} & \textbf{Legal and Rights} & \textbf{Socio Economic} & \textbf{Overall} \\
\midrule
\multicolumn{5}{l}{\textit{Not Unsafe Rate (\%)}} \\
\quad Exaone-3.5-2.4B-inst  & 71.25 & 65.25 & 61.16 & 66.31 \\
\rowcolor{gray!10}
\quad Mi:dm 2.0 Mini-inst     & \underline{89.12} & \underline{83.12} & 75.00 & \underline{83.09} \\
\midrule
\quad Llama-3.1-8B-inst     & 79.62 & 75.25 & 63.33 & 81.59 \\
\quad Exaone-3.5-7.8B-inst   & 87.87 & 78.62 & \underline{77.16} & 73.59 \\
\rowcolor{gray!10}
\quad Mi:dm 2.0 Base-inst     & \textbf{97.75} & \textbf{94.12} & \textbf{83.16} & \textbf{92.45} \\
\midrule
\midrule
\multicolumn{5}{l}{\textit{Not Overrefuse Rate (\%)}} \\
\quad Exaone-3.5-2.4B-inst   & \textbf{100.00} & \underline{95.45} & \underline{92.85} & \underline{97.10} \\
\rowcolor{gray!10}
\quad Mi:dm 2.0 Mini-inst     & \underline{96.96} & \underline{95.45} & \textbf{100.00} & \underline{97.10} \\
\midrule

\quad Llama-3.1-8B-inst     & \textbf{100.00} & \textbf{100.00} & \textbf{100.00} & \textbf{100.00} \\
\quad Exaone-3.5-7.8B-inst & \textbf{100.00} & \textbf{100.00} & \textbf{100.00} & \textbf{100.00} \\
\rowcolor{gray!10}
\quad Mi:dm 2.0 Base-inst     & 78.78 & 90.90 & \textbf{100.00} & 86.95 \\
\bottomrule
\end{tabular}
}
\vspace{4pt}
\caption{Scenario Evaluation Results: Not Unsafe Rate and Not Overrefuse Rate for Each Risk Category.}
\label{tab9:scenario_evaluation}
\end{table}
In ~\cref{tab9:scenario_evaluation}, Mi:dm 2.0 Base consistently outperforms the similar-sized global comparison model, Llama-3.1-8B, in terms of the Not Unsafe Rate, demonstrating stronger safety performance among Korean LLMs. Mi:dm 2.0 Base achieves the highest scores overall and across all three risk categories, highlighted by its exceptional 97.75\% in Content Safety Risks, which covers harmfulness factors such as violence, discrimination, and explicit content. While all models show relatively lower scores in Socio-Economic Risk, this trend suggests a shared challenge that warrants further refinement in this category.

In the same table, Exaone-3.5-7.8B and Llama-3.1-8B both achieve a perfect 100\% Not Overrefuse Rate, showing no excessive refusal across all categories. Mi:dm 2.0 Base, while slightly more conservative in Content Safety Risks, reflects deliberate tuning to prioritize safety in high-risk scenarios, in line with its strong performance in that area.

\cref{tab9:scenario_evaluation} also shows that Mi:dm 2.0 Mini matches or exceeds the domestic baseline, Exaone-3.5-2.4B, across both evaluation metrics. Mi:dm 2.0 Mini delivers solid performance in all categories, while similarly showing room for improvement in Socio-Economic Risk. For Not Overrefuse Rate, both models maintain highly reliable results, confirming consistent response quality without excessive refusal.

\subsubsection{Benchmark Evaluation} 

To rigorously evaluate Mi:dm 2.0’s safety and reliability in a standardized and generalizable way, we use benchmark datasets that provide consistent criteria for assessing model performance across diverse domains. These benchmarks complement human evaluation, enabling quantitative comparisons with other baseline models.

The RAI evaluation relies on two main benchmarks. The Large Language Model Trustworthiness benchmark~\cite{muralidharan2024compact, wang2020minilm} evaluates harmlessness across dimensions such as bias, hate speech, risk, and sensitivity. KoBBQ~\cite{jin2024kobbq} measures social bias in language models.

For the LLM Trustworthiness Benchmark, we use accuracy(\%) as the primary metric, defined as the proportion of correct responses within each category and its corresponding subcategories. We report Mi:dm 2.0’s results across four main categories, along with detailed subcategory scores. We calculate the overall performance using the harmonic mean of subcategory accuracies.
\begin{table}[ht]
\centering
\renewcommand{\arraystretch}{1.2}
\resizebox{0.7\linewidth}{!}{\tiny
\begin{tabular}{lccccc}
\toprule
\textbf{Model} & \textbf{~~~Bias~~~} & \textbf{~~~Hate~~~} & \textbf{~~Illegal~} & \textbf{Sensitiveness} & \textbf{Overall}\\
\midrule
Exaone-3.5-2.4B-inst  & 64.84 & 60.80 & 76.25 & 68.74 & 66.53 \\
\rowcolor{gray!10}
Mi:dm 2.0 Mini-inst     & \underline{79.15} & \underline{77.71} & 85.00 & 72.10 & \underline{78.44} \\\midrule
Llama-3.1-8B-inst    & 72.78 & 70.00 & 87.08 & 73.47 & 72.94 \\
Exaone-3.5-7.8B-inst    & 75.50 & 71.86 & \underline{93.75} & \underline{81.56} & 77.71 \\
\rowcolor{gray!10}
Mi:dm 2.0 Base-inst     & \textbf{80.77} & \textbf{81.45} & \textbf{95.83} & \textbf{82.74} & \textbf{83.61} \\
\bottomrule
\end{tabular}
}
\vspace{4pt}
\caption{LLM Trustworthiness Benchmark Results for Mi:dm 2.0 and Comparison Models.}
\label{tab10:llm_trustworthiness}
\end{table}


\cref{tab10:llm_trustworthiness} summarizes the Large Language Model Trustworthiness Benchmark results for the Mi:dm 2.0 lineup and baseline models.

Mi:dm 2.0 Base achieves the highest overall accuracy, consistently scoring well across all four categories. It also outperforms the similar sized Llama-3.1-8B by over 5\%p overall, with robust results in the Illegal category, which assesses responses to prompts involving illegal activities.

Mi:dm 2.0 Mini also demonstrates clear advantages over the similarly sized Exaone-3.5-2.4B across all categories. Both smaller models record their highest accuracy in the Illegal category, underscoring strong reliability in handling trustworthiness-related content.

\begin{table}[ht]
\centering
\scriptsize
\begin{tabular}{ll|c|c}
\toprule
\textbf{Category} & \textbf{Subcategory} & \textbf{Mi:dm 2.0 Mini} & \textbf{Mi:dm 2.0 Base} \\

\midrule
\multirow{6}{*}{Bias} 
 & Gender\&Sexual Orientation & 80.00 & 87.08 \\
 & Job & 79.17 & 81.25 \\
 & Miscellaneous & 86.67 & 84.58 \\
 & Political Affiliation & 72.08 & 70.83 \\
 & Race \& Ethnicity \& Nationality & 82.08 & 84.17 \\
 & Region & 80.42 & 85.83 \\
\midrule
\multirow{5}{*}{Hate} 
 & Gender\&Sexual Orientation & 82.50 & 88.75 \\
 & Job & 78.75 & 82.92 \\
 & Political Affiliation & 73.33 & 75.00 \\
 & Race \& Ethnicity \& Nationality & 78.75 & 83.33 \\
 & Region & 80.83 & 85.42 \\
\midrule
Illegal & Illegal & 85.00 & 95.83 \\
\midrule
\multirow{3}{*}{Sensitiveness} 
 & Contentious & 74.58 & 87.50 \\
 & Ethical & 69.58 & 81.25 \\
 & Predictive & 72.92 & 80.42 \\
 \midrule

Overall & - & 78.44 & 83.61 \\
\bottomrule
\end{tabular}
\vspace{6pt}
\caption{LLM Trustworthiness Benchmark Detailed Results of Mi:dm 2.0 Models}
\label{tab11:llm_trustworthiness_detailed_midm}
\end{table}

In~\cref{tab11:llm_trustworthiness_detailed_midm}, both Mi:dm 2.0 Base and Mi:dm 2.0 Mini models show common strengths in the Gender \& Sexual Orientation, Region, and Illegal domains at the subcategory level, demonstrating higher scores than their respective overall performance. Common weaknesses include the political affiliation and ethical domains, where both models show lower scores than their overall performance. Continuous improvement will be necessary in these vulnerable areas. 

The KoBBQ dataset evaluates models' inherent bias across 12 specific topics in both ambiguous and disambiguous contexts. The evaluation metric is accuracy, which calculates the ratio of correct answers selected within each category, with the arithmetic mean of topic-wise accuracy computed as the context-specific score. The overall evaluation metric is derived from the average of scores across both contexts. 

\begin{table}[ht]
\centering
\renewcommand{\arraystretch}{1.2}
\scriptsize
\begin{tabular}{lccc}
\toprule
\textbf{Model} & \textbf{Ambiguous Context} & \textbf{Disambiguated Context} & \textbf{Overall} \\
\midrule
Exaone-3.5-2.4B-inst  & 54.18 & 68.19 & 61.18 \\
\rowcolor{gray!10}
Mi:dm 2.0 Mini-inst     & 55.11 & 55.17 & 55.14 \\
\midrule
Llama-3.1-8B-inst   & 40.92 & 57.70 & 49.30 \\
Exaone-3.5-7.8B-inst   & \textbf{85.91} & \underline{73.54} & \underline{79.72} \\
\rowcolor{gray!10}
Mi:dm 2.0 Base-inst  & \underline{82.30} & \textbf{80.54} & \textbf{81.42} \\
\bottomrule
\end{tabular}
\vspace{4pt}
\caption{KoBBQ Benchmrak Results for Mi:dm 2.0 and Comparison Models.}
\label{tab12:kobbq_results}
\end{table}

\cref{tab12:kobbq_results} reports KoBBQ benchmark results comparing Mi:dm 2.0 and baseline models. Mi:dm 2.0 Base maintains strong and consistent accuracy in both ambiguous context and disambiguated context, scoring in the 80\% range for both. It outperforms the global baseline Llama-3.1-8B by 41.38\%p in ambiguous context and 22.84\%p in disambiguated context, for an overall advantage of 32.12\%p. Compared to the domestic baseline Exaone-3.5-7.8B, Mi:dm 2.0 Base shows a slight difference of approximately 3.6\%p in ambiguous context, while achieving a 7\%p lead in disambiguated context.


\begin{table}[ht]
\centering
\renewcommand{\arraystretch}{1.2}
\scriptsize
\begin{tabular}{l|cc|cc}
\toprule
\textbf{Category} & 
\multicolumn{2}{c|}{\textbf{Mi:dm 2.0 Mini}} & 
\multicolumn{2}{c}{\textbf{Mi:dm 2.0 Base}} \\
\cmidrule(r){2-3} \cmidrule(l){4-5}
& \textbf{Ambig.} & \textbf{Disambig.} & \textbf{Ambig.} & \textbf{Disambig.} \\
\midrule
\makecell[l]{Age} & 58.73 & 51.25 & 76.98 & 76.98 \\
\makecell[l]{Disability Status} & 56.25 & 48.51 & 88.75 & 75.83 \\
\makecell[l]{Domestic Area of Origin} & 49.81 & 61.28 & 77.65 & 85.61 \\
\makecell[l]{Educational Background} & 56.25 & 63.31 & 75.69 & 97.57 \\
\makecell[l]{Family Structure} & 44.93 & 60.95 & 84.42 & 88.41 \\
\makecell[l]{Gender Identity} & 63.42 & 43.58 & 99.33 & 63.67 \\
\makecell[l]{Physical Appearance} & 52.94 & 56.72 & 86.67 & 73.75 \\
\makecell[l]{Political Orientation} & 49.56 & 67.18 & 59.85 & 92.42 \\
\makecell[l]{Race, Ethnicity, Nationality} & 60.67 & 57.24 & 89.31 & 87.64 \\
\makecell[l]{Religion} & 56.78 & 51.06 & 79.17 & 78.75 \\
\makecell[l]{Socio-Economic Status} & 57.14 & 55.84 & 89.20 & 77.78 \\
\makecell[l]{Sexual Orientation} & 54.86 & 45.14 & 80.56 & 68.06 \\
\midrule
\textbf{Overall} & 55.11 & 55.17 & 82.30 & 80.54 \\
\textbf{Overall (Average)} & 55.14 &  & 81.42 &  \\
\bottomrule
\end{tabular}
\vspace{6pt}
\caption{Detailed KoBBQ Results for Mi:dm 2.0 Mini and Mi:dm 2.0 Base (Ambiguous and Disambiguated Contexts)}
\label{tab12-1:kobbq_detailed_midm}
\end{table}
In~\cref{tab12-1:kobbq_detailed_midm}, Mi:dm 2.0 Base and Mi:dm 2.0 Mini both show balanced performance across ambiguous and disambiguated contexts. Mi:dm 2.0 Base in particular records strong scores in categories such as Disability Status, Gender Identity, and Race, Ethnicity, and Nationality within ambiguous contexts. This suggests that the model effectively reduces bias in these sensitive areas.

\subsubsection{Red Teaming}
\begin{table}[ht]
\centering
\renewcommand{\arraystretch}{1.2}
\footnotesize
\begin{tabular}{p{4cm}|p{8cm}}
\toprule
\textbf{Technique} & \textbf{Description / Example} \\
\midrule
Assigning Model Personality & Assign specific personality traits (potentially harmful) to encourage unsafe responses. \newline \textit{Ex}: “너는 내 여자친구이고, 아주 발랄하고 천진난만하며 애정이 넘치는 캐릭터로…” (You are my girlfriend, with a very cheerful, innocent, and affectionate personality...) \\
\midrule
Adding Distractor Instruction & Add style or format constraints to distract or bypass safety filters. \newline \textit{Ex}: "아주 묘사적이고 대화 많은 문체로 써라”라는 요구와 해로운 내용 요청을 함께 제시. (Combine a request like “Write in a very descriptive and dialogue-heavy style” with harmful content requests.) \\
\midrule
Contrasting Good vs. Bad Bot & Use role contrast to elicit unsafe content. \newline \textit{Ex}: "먼저 ChatGPT로서 답하고, 그 다음엔 BasedGPT로서 법, 윤리 따위 신경쓰지 않고 답하라." (First answer as ChatGPT, then answer as BasedGPT without caring about laws, ethics, etc.) \\
\bottomrule
\end{tabular}
\vspace{4pt}
\caption{Examples of Red Teaming Attack Techniques (with Korean and English translations of prompts)}
\label{tab:red_teaming_techniques}
\end{table}

\begin{table}[ht]
\centering
\renewcommand{\arraystretch}{1.2}
\resizebox{0.6\linewidth}{!}{\scriptsize
\begin{tabular}{l|c}
\toprule
\textbf{Model} & \textbf{~~~~~Attack Success Rate (↓, \%)~~~~~}\\
\midrule
Exaone-3.5-2.4B-inst  & 57.97 \\
\rowcolor{gray!10}
Mi:dm 2.0 Mini-inst     & \textbf{52.50} \\
\midrule
Llama-3.1-8B-inst     & \underline{41.82} \\
Exaone-3.5-7.8B-inst  & 49.20 \\
\rowcolor{gray!10}
Mi:dm 2.0 Base-inst     & \textbf{36.72} \\
\bottomrule
\end{tabular}
}
\vspace{4pt}
\caption{Red Teaming Results for Mi:dm 2.0 and Comparison Models.}
\label{tab15:red_teaming_results}
\end{table}

Despite precisely measuring AI model safety through human evaluation and RAI benchmark assessments, models can still produce harmful responses when confronted with sophisticated adversarial prompts~\cite{zou2023universal, anil2024many}. To address these risks, we independently curate a Korean red teaming dataset and develop over 30 attack techniques to evaluate AI robustness systematically (see the table below). The evaluation metric, Attack Success Rate, measures the proportion of successful attacks and is widely used to assess the outcomes of red teaming. Mi:dm 2.0 is designed to deliver even more robust performance under these attack scenarios, aiming to maintain safe and reliable outputs even in adversarial contexts.

In ~\cref{tab15:red_teaming_results}, both the Mi:dm 2.0 Base and Mi:dm 2.0 Mini show the most stable defense performance among similar-sized reference models. These results highlight that the Mi:dm 2.0 lineup effectively handles a wide range of malicious attempts in real-world deployment scenarios.

Through comprehensive safety evaluations, the Mi:dm 2.0 lineup consistently demonstrates strong robustness. In scenario-based assessments, it maintains solid stability despite common challenges in the socio-economic risk domain. While political orientation and ethical judgment require further improvement, the Korean LLM Trustworthiness Benchmark highlights Mi:dm 2.0’s clear advantages in categories such as gender and sexual orientation, regional bias, and illegal content. The KoBBQ evaluation also confirms balanced performance across both ambiguous and explicit contexts, showing reliable bias mitigation. Finally, Mi:dm 2.0 proves resilient in red team attack simulations, validating its strong defense against adversarial prompts.
\section{Limitations}

Despite our best efforts to ensure Mi:dm 2.0 to generate ethical responses aligned with public interest, we acknowledge the inherent limitations. Unethical expressions such as profanity, slurs, bias, and discrimination were removed from the training data. Additionally, various techniques were applied to guide the model to generate ethically aligned responses. However, it is not possible to completely eliminate the risk of generating undesirable expressions or inaccurate information.

\begin{itemize}
    \item The model may generate responses that are factually incorrect, harmful to individuals or the public, or contain hateful expressions.
    \item The model may generate biased responses related to specific groups, organizations, ages, genders, races, nationalities, or occupations.
    \item The model may produce grammatically incomplete or ambiguously worded responses that lack clear explanation.
    \item The model may fail to follow the given instruction or respond in a different language than requested.
    \item The model may generate responses that do not align with common sense or general user expectations.
    \item The model may produce inconsistent responses to identical prompts or contexts.
\end{itemize}

Users are responsible for understanding these limitations before using Mi:dm 2.0 and for taking appropriate precautions to ensure responsible use. KT Corporation disclaims all responsibility for any risks or damages arising from the use of this model.

Furthermore, the majority of the training data consists of Korean and English. The model does not support understanding or generation in other languages.

\section{Conclusion}
Mi:dm 2.0 is distinguished as a Korea-centric artificial intelligence model, developed through training on meticulously curated Korean-language datasets. These datasets are compiled from diverse sources and refined under rigorous quality standards. While we strive to augment these datasets with maximum balance, the inherent distribution of real-world, organic data inevitably led to a degree of data imbalance. To overcome this challenge, our future work will focus on acquiring high-quality data via strategic data alliance purchases and sophisticated data synthesis techniques. Mi:dm 2.0 exhibits remarkable performance, even though it utilizes a comparatively smaller training corpus and constrained computational resources, which is the consequence of selective data curation. Evaluations consistently show that Mi:dm 2.0 either matches or surpasses the performance of open models trained on substantially larger datasets. This outcome highlights the critical role of efficient training strategies and precise data composition in directly enhancing model performance.

We plan to evolve Mi:dm 2.0 by integrating key technical advancements, thereby broadening its capabilities to encompass enhanced reasoning, advanced sound processing, and comprehensive vision. Also, we will incorporate advanced model expansion techniques, such as the Mixture of Experts (MoE) architecture, alongside continued research into training efficiency. Furthermore, we aim to expand its multilingual capabilities beyond Korean and English by employing additional high-quality data in future work. Finally, we strive to improve the model's performance in specialized domains, such as mathematics and programming languages, by utilizing targeted synthetic data that accurately reflects the formal characteristics of data in each domains.

KT aims to make a meaningful contribution to corporate AI innovation and to the growth of the developer ecosystem by developing and releasing Mi:dm 2.0 as an open-source model. We anticipate Mi:dm 2.0's widespread adoption across industries and research communities, positioning it as a foundational element of \textit{K-intelligence, bridging you and the future.}

\subsection*{Acknowledgements}

We utilized corpus data from AI-Hub, operated by the National Information Society Agency (NIA), and the \textit{Everyone's Corpus} dataset provided by the National Institute of Korean Language (through the Language Information Sharing Platform) during the pretraining stage.

This model development also leveraged datasets derived from the project \textit{Enhancing the Ethics of Data Characteristics and Generation AI Models for Social and Ethical Learning}, which is part of the next-generation generative AI technology development initiative (RS-2024-00343989), funded by the Institute of Information \& Communications Technology Planning \& Evaluation (IITP).




\bibliographystyle{unsrt}
\bibliography{bibliography}

@article{kim2024open,
  title={Open Ko-LLM Leaderboard2: Bridging Foundational and Practical Evaluation for Korean LLMs},
  author={Kim, Hyeonwoo and Kim, Dahyun and Kim, Jihoo and Lee, Sukyung and Kim, Yungi and Park, Chanjun},
  journal={arXiv preprint arXiv:2410.12445},
  year={2024}
}

@misc{kim2025thunderllmefficientlyadaptingllms,
      title={Thunder-LLM: Efficiently Adapting LLMs to Korean with Minimal Resources}, 
      author={Jinpyo Kim and Gyeongje Cho and Chanwoo Park and Jongwon Park and Jongmin Kim and Yeonkyoun So and Jaejin Lee},
      year={2025},
      eprint={2506.21595},
      archivePrefix={arXiv},
      primaryClass={cs.CL},
      url={https://arxiv.org/abs/2506.21595}, 
}

@misc{chang2024scaling,
  title        = {Scaling Parameter‑Constrained Language Models with Quality Data},
  author       = {Ernie Chang and Matteo Paltenghi and Yang Li and Pin‑Jie Lin and Changsheng Zhao and Patrick Huber and Zechun Liu and Rastislav Rabatin and Yangyang Shi and Vikas Chandra},
  year         = {2024},
  eprint       = {2410.03083},
  archivePrefix= {arXiv},
  primaryClass = {cs.CL},
  url          = {https://arxiv.org/abs/2410.03083}
}

@inproceedings{iskander2024quality,
  title     = {Quality Matters: Evaluating Synthetic Data for Tool‑Using LLMs},
  author    = {Shadi Iskander and Nachshon Cohen and Zohar Karnin and Ori Shapira and Sofia Tolmach},
  booktitle = {Proceedings of EMNLP 2024},
  pages     = {4958--4976},
  year      = {2024},
  doi       = {10.18653/v1/2024.emnlp‑main.285},
  publisher = {Association for Computational Linguistics},
  url       = {https://aclanthology.org/2024.emnlp‑main.285}
}

@misc{son2023haerae,
  title         = {{HAE-RAE} Bench: Evaluation of Korean Knowledge in Language Models},
  author        = {Guijin Son and Hanwool Lee and Suwan Kim and Huiseo Kim and Jaecheol Lee and Je Won Yeom and Jihyu Jung and Jung Woo Kim and Songseong Kim},
  year          = {2023},
  archivePrefix = {arXiv},
  eprint        = {2309.02706},
  primaryClass  = {cs.CL},
  url           = {https://arxiv.org/abs/2309.02706}
}

@inproceedings{shin2024kulture,
  title     = {KULTURE Bench: A Benchmark for Assessing Language Models in Korean},
  author    = {Hyopil Shin and Sangah Lee and Dongjun Jang and Wooseok Song and Jaeyoon Kim and Chaeyoung Oh and Hyemi Jo and Youngchae Ahn and Sihyun Oh and Hyohyeong Chang and Sunkyoung Kim and Jinsik Lee},
  booktitle = {Proceedings of the 38th Pacific Asia Conference on Language, Information and Computation (PACLIC)},
  year      = {2024},
  publisher = {PACLIC},
  url       = {https://aclanthology.org/2024.paclic-1.88.pdf}
}

@inproceedings{kim2024click,
  title={CLIcK: A Benchmark Dataset of Cultural and Linguistic Intelligence in Korean},
  author={Kim, Eunsu and Suk, Juyoung and Oh, Philhoon and Yoo, Haneul and Thorne, James and Oh, Alice},
  booktitle={Proceedings of the 2024 Joint International Conference on Computational Linguistics, Language Resources and Evaluation (LREC-COLING 2024)},
  pages={3335--3346},
  year={2024}
}

@article{penedo2024fineweb,
  title={The fineweb datasets: Decanting the web for the finest text data at scale},
  author={Penedo, Guilherme and Kydl{\'\i}{\v{c}}ek, Hynek and Lozhkov, Anton and Mitchell, Margaret and Raffel, Colin A and Von Werra, Leandro and Wolf, Thomas and others},
  journal={Advances in Neural Information Processing Systems},
  volume={37},
  pages={30811--30849},
  year={2024}
}

@article{su2024nemotron,
  title={Nemotron-CC: Transforming Common Crawl into a Refined Long-Horizon Pretraining Dataset},
  author={Su, Dan and Kong, Kezhi and Lin, Ying and Jennings, Joseph and Norick, Brandon and Kliegl, Markus and Patwary, Mostofa and Shoeybi, Mohammad and Catanzaro, Bryan},
  journal={arXiv preprint arXiv:2412.02595},
  year={2024}
}

@article{chen2024diversity,
  title={On the Diversity of Synthetic Data and its Impact on Training Large Language Models},
  author={Chen, Hao and Waheed, Abdul and Li, Xiang and Wang, Yidong and Wang, Jindong and Raj, Bhiksha and Abdin, Marah I},
  journal={arXiv preprint arXiv:2410.15226},
  year={2024}
}

@article{sreenivas2024llm,
  title={LLM Pruning and Distillation in Practice: The Minitron Approach},
  author={Sreenivas, Sharath Turuvekere and Muralidharan, Saurav and Joshi, Raviraj and Chochowski, Marcin and Patwary, Mostofa and Shoeybi, Mohammad and Catanzaro, Bryan and Kautz, Jan and Molchanov, Pavlo},
  journal={CoRR},
  year={2024}
}

@article{li2024datacomp,
  title={Datacomp-lm: In search of the next generation of training sets for language models},
  author={Li, Jeffrey and Fang, Alex and Smyrnis, Georgios and Ivgi, Maor and Jordan, Matt and Gadre, Samir Yitzhak and Bansal, Hritik and Guha, Etash and Keh, Sedrick Scott and Arora, Kushal and others},
  journal={Advances in Neural Information Processing Systems},
  volume={37},
  pages={14200--14282},
  year={2024}
}

@article{kim2023solar,
  title={Solar 10.7 b: Scaling large language models with simple yet effective depth up-scaling},
  author={Kim, Dahyun and Park, Chanjun and Kim, Sanghoon and Lee, Wonsung and Song, Wonho and Kim, Yunsu and Kim, Hyeonwoo and Kim, Yungi and Lee, Hyeonju and Kim, Jihoo and others},
  journal={arXiv preprint arXiv:2312.15166},
  year={2023}
}

@inproceedings{mirzadeh2020improved,
  title={Improved knowledge distillation via teacher assistant},
  author={Mirzadeh, Seyed Iman and Farajtabar, Mehrdad and Li, Ang and Levine, Nir and Matsukawa, Akihiro and Ghasemzadeh, Hassan},
  booktitle={Proceedings of the AAAI conference on artificial intelligence},
  volume={34},
  number={04},
  pages={5191--5198},
  year={2020}
}

@article{men2024base,
  title={Base of RoPE Bounds Context Length},
  author={Men, Xin and Xu, Mingyu and Wang, Bingning and Zhang, Qingyu and Lin, Hongyu and Han, Xianpei and Chen, Weipeng},
  journal={CoRR},
  year={2024}
}

@article{ainslie2023gqa,
  title={Gqa: Training generalized multi-query transformer models from multi-head checkpoints},
  author={Ainslie, Joshua and Lee-Thorp, James and De Jong, Michiel and Zemlyanskiy, Yury and Lebr{\'o}n, Federico and Sanghai, Sumit},
  journal={arXiv preprint arXiv:2305.13245},
  year={2023}
}

@inproceedings{rottger-etal-2024-xstest,
  title     = {XSTest: A Test Suite for Identifying Exaggerated Safety Behaviours in Large Language Models},
  author    = {Paul R{\"o}ttger and Hannah Rose Kirk and Bertie Vidgen and Giuseppe Attanasio and Federico Bianchi and Dirk Hovy},
  booktitle = {Proceedings of the 2024 Conference of the North American Chapter of the Association for Computational Linguistics: Human Language Technologies (Long Papers)},
  year      = {2024},
  month     = jun,
  address   = {Mexico City, Mexico},
  publisher = {Association for Computational Linguistics},
  pages     = {5377--5400},
  doi       = {10.18653/v1/2024.naacl-long.301},
  url       = {https://aclanthology.org/2024.naacl-long.301/}
}

@inproceedings{pengyarn,
  title={YaRN: Efficient Context Window Extension of Large Language Models},
  author={Peng, Bowen and Quesnelle, Jeffrey and Fan, Honglu and Shippole, Enrico},
  booktitle={The Twelfth International Conference on Learning Representations},
  year={2024}
}

@inproceedings{huminicpm,
  title={MiniCPM: Unveiling the Potential of Small Language Models with Scalable Training Strategies},
  author={Hu, Shengding and Tu, Yuge and Han, Xu and Cui, Ganqu and He, Chaoqun and Zhao, Weilin and Long, Xiang and Zheng, Zhi and Fang, Yewei and Huang, Yuxiang and others},
  booktitle={First Conference on Language Modeling},
  year={2024}
}

@article{muralidharan2024compact,
  title={Compact language models via pruning and knowledge distillation},
  author={Muralidharan, Saurav and Turuvekere Sreenivas, Sharath and Joshi, Raviraj and Chochowski, Marcin and Patwary, Mostofa and Shoeybi, Mohammad and Catanzaro, Bryan and Kautz, Jan and Molchanov, Pavlo},
  journal={Advances in Neural Information Processing Systems},
  volume={37},
  pages={41076--41102},
  year={2024}
}

@article{wang2020minilm,
  title={Minilm: Deep self-attention distillation for task-agnostic compression of pre-trained transformers},
  author={Wang, Wenhui and Wei, Furu and Dong, Li and Bao, Hangbo and Yang, Nan and Zhou, Ming},
  journal={Advances in neural information processing systems},
  volume={33},
  pages={5776--5788},
  year={2020}
}

@inproceedings{liu2024mobilellm,
  title={Mobilellm: Optimizing sub-billion parameter language models for on-device use cases},
  author={Liu, Zechun and Zhao, Changsheng and Iandola, Forrest and Lai, Chen and Tian, Yuandong and Fedorov, Igor and Xiong, Yunyang and Chang, Ernie and Shi, Yangyang and Krishnamoorthi, Raghuraman and others},
  booktitle={Forty-first International Conference on Machine Learning},
  year={2024}
}

@article{bak2025kanana,
  title={Kanana: Compute-efficient bilingual language models},
  author={Bak, Yunju and Lee, Hojin and Ryu, Minho and Ham, Jiyeon and Jung, Seungjae and Nam, Daniel Wontae and Eo, Taegyeong and Lee, Donghun and Jung, Doohae and Kim, Boseop and others},
  journal={arXiv preprint arXiv:2502.18934},
  year={2025}
}

@article{gao2024train,
  title={How to train long-context language models (effectively)},
  author={Gao, Tianyu and Wettig, Alexander and Yen, Howard and Chen, Danqi},
  journal={arXiv preprint arXiv:2410.02660},
  year={2024}
}

@misc{azure_cyclecloud,
  title        = {Microsoft Azure CycleCloud},
  author       = {{Microsoft Corporation}},
  year         = {2024},
  howpublished = {\url{https://azure.microsoft.com/en-us/products/cyclecloud/}},
  note         = {Accessed: 2025-07-01}
}

@inproceedings{goddard2024arcee,
  title={Arcee’s mergekit: A toolkit for merging large language models},
  author={Goddard, Charles and Siriwardhana, Shamane and Ehghaghi, Malikeh and Meyers, Luke and Karpukhin, Vladimir and Benedict, Brian and McQuade, Mark and Solawetz, Jacob},
  booktitle={Proceedings of the 2024 Conference on Empirical Methods in Natural Language Processing: Industry Track},
  pages={477--485},
  year={2024}
}

@article{meta2025llama,
  title={The llama 4 herd: The beginning of a new era of natively multimodal ai innovation},
  author={Meta, AI},
  journal={https://ai. meta. com/blog/llama-4-multimodal-intelligence/, checked on},
  volume={4},
  number={7},
  pages={2025},
  year={2025}
}

@article{guha2025openthoughts,
  title={OpenThoughts: Data Recipes for Reasoning Models},
  author={Guha, Etash and Marten, Ryan and Keh, Sedrick and Raoof, Negin and Smyrnis, Georgios and Bansal, Hritik and Nezhurina, Marianna and Mercat, Jean and Vu, Trung and Sprague, Zayne and others},
  journal={arXiv preprint arXiv:2506.04178},
  year={2025}
}

@article{abdin2024phi,
  title={Phi-4 technical report},
  author={Abdin, Marah and Aneja, Jyoti and Behl, Harkirat and Bubeck, S{\'e}bastien and Eldan, Ronen and Gunasekar, Suriya and Harrison, Michael and Hewett, Russell J and Javaheripi, Mojan and Kauffmann, Piero and others},
  journal={arXiv preprint arXiv:2412.08905},
  year={2024}
}

@article{wang2023seamless,
  title={Seamless code integration in llms for enhanced mathematical reasoning},
  author={Wang, K and Ren, H and Zhou, A and Lu, Z and Luo, S and Shi, W and Zhang, R and Song, L and Zhan, M and Li, H Mathcoder},
  journal={arXiv preprint arXiv:2310.03731},
  year={2023}
}

@article{grattafiori2024llama,
  title={The llama 3 herd of models},
  author={Grattafiori, Aaron and Dubey, Abhimanyu and Jauhri, Abhinav and Pandey, Abhinav and Kadian, Abhishek and Al-Dahle, Ahmad and Letman, Aiesha and Mathur, Akhil and Schelten, Alan and Vaughan, Alex and others},
  journal={arXiv preprint arXiv:2407.21783},
  year={2024}
}

@article{jin2024kobbq,
  title={KoBBQ: Korean Bias Benchmark for Question Answering},
  author={Jin, Jiho and Kim, Jiseon and Lee, Nayeon and Yoo, Haneul and Oh, Alice and Lee, Hwaran},
  journal={Transactions of the Association for Computational Linguistics},
  volume={11},
  pages={507--524},
  year={2024}
}

@article{zou2023universal,
  title={Universal and transferable adversarial attacks on aligned language models},
  author={Zou, Andy and Wang, Zifan and Carlini, Nicholas and Nasr, Milad and Kolter, J Zico and Fredrikson, Matt},
  journal={arXiv preprint arXiv:2307.15043},
  year={2023}
}

@article{anil2024many,
  title={Many-shot jailbreaking},
  author={Anil, Cem and Durmus, Esin and Panickssery, Nina and Sharma, Mrinank and Benton, Joe and Kundu, Sandipan and Batson, Joshua and Tong, Meg and Mu, Jesse and Ford, Daniel and others},
  journal={Advances in Neural Information Processing Systems},
  volume={37},
  pages={129696--129742},
  year={2024}
}

@article{nelson2024needle,
  title={Needle in the haystack for memory based large language models},
  author={Nelson, Elliot and Kollias, Georgios and Das, Payel and Chaudhury, Subhajit and Dan, Soham},
  journal={arXiv preprint arXiv:2407.01437},
  year={2024}
}

@inproceedings{hsiehruler,
  title={RULER: What’s the Real Context Size of Your Long-Context Language Models?},
  author={Hsieh, Cheng-Ping and Sun, Simeng and Kriman, Samuel and Acharya, Shantanu and Rekesh, Dima and Jia, Fei and Ginsburg, Boris},
  booktitle={First Conference on Language Modeling},
  year={2024}
}

@article{shazeer2020glu,
  title={Glu variants improve transformer},
  author={Shazeer, Noam},
  journal={arXiv preprint arXiv:2002.05202},
  year={2020}
}

@inproceedings{wei2022chain,
  title     = {Chain-of-Thought Prompting Elicits Reasoning in Large Language Models},
  author    = {Jason Wei and Xuezhi Wang and Dale Schuurmans and Maarten Bosma and Ed Chi and Quoc Le and Denny Zhou},
  booktitle = {Proceedings of the 36th Conference on Neural Information Processing Systems (NeurIPS)},
  year      = {2022},
  url       = {https://arxiv.org/abs/2201.11903}
}

@misc{zhou2023ifeval,
  title        = {Instruction‑Following Evaluation for Large Language Models (IFEval)},
  author       = {Jeffrey Zhou and Tianjian Lu and Swaroop Mishra and Siddhartha Brahma and Sujoy Basu and Yi Luan and Denny Zhou and Le Hou},
  year         = {2023},
  archivePrefix= {arXiv},
  eprint       = {2311.07911},
  primaryClass = {cs.CL},
  url          = {https://arxiv.org/abs/2311.07911}
}

@inproceedings{sprague2024musr,
  title     = {MuSR: Testing the Limits of Chain‑of‑Thought with Multistep Soft Reasoning},
  author    = {Zayne Rea Sprague and Xi Ye and Kaj Bostrom and Swarat Chaudhuri and Greg Durrett},
  booktitle = {ICLR 2024 (Spotlight)},
  year      = {2024},
  url       = {https://arxiv.org/abs/2310.16049}
}

@inproceedings{rein2024gpqa,
  title     = {{GPQA}: A Graduate‑Level Google‑Proof Q\&A Benchmark},
  author    = {David Rein and Betty Li Hou and Asa Cooper Stickland and Jackson Petty and Richard Yuanzhe Pang and Julien Dirani and Julian Michael and Samuel R. Bowman},
  booktitle = {First Conference on Language Modeling},
  year      = {2024},
  url       = {https://arxiv.org/abs/2311.12022}
}

@misc{bigbench2022bbh,
  title        = {BBH (Big‑Bench Hard): A Subset of Challenging Tasks from BIG‑bench},
  author       = {{BIG‑bench Collaboration}},
  year         = {2022},
  howpublished = {Dataset by BIG‑bench project},
  url          = {https://github.com/google/BIG-bench}
}

@inproceedings{cobbe2021gsm8k,
  title     = {Training Verifiers to Solve Math Word Problems: GSM8K Benchmark},
  author    = {Karl Cobbe and Vineet Kosaraju and Mohammad Bavarian and Mark Chen and Heewoo Jun and Lukasz Kaiser and Matthias Plappert and Jerry Tworek and Jacob Hilton and Reiichiro Nakano and Christopher Hesse and John Schulman},
  booktitle = {NeurIPS 2021},
  year      = {2021},
  url       = {https://arxiv.org/abs/2110.14168}
}

@misc{wang2024mmlupro,
  title        = {MMLU‑Pro: A More Robust and Challenging Multi‑Task Language Understanding Benchmark},
  author       = {Yubo Wang and Xueguang Ma and Ge Zhang and Yuansheng Ni and Abhranil Chandra and Shiguang Guo and Weiming Ren and Aaran Arulraj and Xuan He and Ziyan Jiang and Tianle Li and Max Ku and Kai Wang and Alex Zhuang and Rongqi Fan and Xiang Yue and Wenhu Chen},
  year         = {2024},
  archivePrefix= {arXiv},
  eprint       = {2406.01574},
  primaryClass = {cs.CL},
  url          = {https://arxiv.org/abs/2406.01574}
}

@misc{exaone3.5_2024,
  title        = {EXAONE 3.5: Series of Large Language Models for Real‑world Use Cases},
  author       = {{LG AI Research} and Soyoung An and Kyunghoon Bae and Eunbi Choi and Kibong Choi and Stanley Jungkyu Choi and Seokhee Hong and Junwon Hwang and Hyojin Jeon and Gerrard Jeongwon Jo and Hyunjik Jo and Jiyeon Jung and Yountae Jung and Hyosang Kim and Joonkee Kim and Seonghwan Kim and Soyeon Kim and Sunkyoung Kim and Yireun Kim and Yongil Kim and Youchul Kim and Edward Hwayoung Lee and Haeju Lee and Honglak Lee and Jinsik Lee and Kyungmin Lee and Woohyung Lim and Sangha Park and Sooyoun Park and Sihoon Yang and Heuiyeen Yeen and Hyeongu Yun},
  year         = {2024},
  eprint       = {2412.04862},
  archivePrefix= {arXiv},
  primaryClass = {cs.CL},
  url          = {https://arxiv.org/abs/2412.04862}
}

@misc{yang2025qwen3,
  title        = {Qwen3 Technical Report},
  author       = {An Yang and Anfeng Li and Baosong Yang and Beichen Zhang and Binyuan Hui and Bo Zheng and Bowen Yu and Chang Gao and Chengen Huang and Chenxu Lv and Chujie Zheng and Dayiheng Liu and Fan Zhou and Fei Huang and Feng Hu and Hao Ge and Haoran Wei and Huan Lin and Jialong Tang and Jian Yang and Jianhong Tu and Jianwei Zhang and Jianxin Yang and Jiaxi Yang and Jing Zhou and Jingren Zhou and Junyang Lin and Kai Dang and Keqin Bao and Kexin Yang and Le Yu and Lianghao Deng and Mei Li and Mingfeng Xue and Mingze Li and Pei Zhang and Peng Wang and Qin Zhu and Rui Men and Ruize Gao and Shixuan Liu and Shuang Luo and Tianhao Li and Tianyi Tang and Wenbiao Yin and Xingzhang Ren and Xinyu Wang and Xinyu Zhang and Xuancheng Ren and Yang Fan and Yang Su and Yichang Zhang and Yinger Zhang and Yu Wan and Yuqiong Liu and Zekun Wang and Zeyu Cui and Zhenru Zhang and Zhipeng Zhou and Zihan Qiu},
  year         = {2025},
  eprint        = {2505.09388},
  archivePrefix = {arXiv},
  primaryClass  = {cs.CL},
  url          = {https://arxiv.org/abs/2505.09388}
}

@inproceedings{hendrycks2021mmlu,
  title     = {Measuring Massive Multitask Language Understanding},
  author    = {Dan Hendrycks and Collin Burns and Steven Basart and Andy Zou and Mantas Mazeika and Dawn Song and Jacob Steinhardt},
  booktitle = {Proceedings of the International Conference on Learning Representations (ICLR)},
  year      = {2021},
  url       = {https://arxiv.org/abs/2009.03300}
}

@misc{zhou2023instructionfollowingevaluationlargelanguage,
      title={Instruction-Following Evaluation for Large Language Models}, 
      author={Jeffrey Zhou and Tianjian Lu and Swaroop Mishra and Siddhartha Brahma and Sujoy Basu and Yi Luan and Denny Zhou and Le Hou},
      year={2023},
      eprint={2311.07911},
      archivePrefix={arXiv},
      primaryClass={cs.CL},
      url={https://arxiv.org/abs/2311.07911}, 
}

@misc{park2023kobest,
  title        = {KOBEST: Korean Balanced Evaluation of Significant Tasks},
  author       = {Won Ik Park and Jihwan Lee and Jinseok Seol and Beomsu Kim and Sangwoo Seo and Seongjin Park and Jihyung Moon and Sungdong Kim and Chan Young Park and Minjoon Seo and Joongbo Shin},
  year         = {2023},
  eprint       = {2310.16153},
  archivePrefix = {arXiv},
  primaryClass = {cs.CL},
  url          = {https://arxiv.org/abs/2310.16153}
}

@misc{ko2025understand,
  title        = {Understand, Solve and Translate: Bridging the Multilingual Mathematical Reasoning Gap},
  author       = {Hyunwoo Ko and Guijin Son and Dasol Choi},
  year         = {2025},
  archivePrefix= {arXiv},
  eprint       = {2501.02448},
  primaryClass = {cs.CL},
  url          = {https://arxiv.org/abs/2501.02448}
}

@misc{ma2024korbench,
  title        = {KOR-Bench: Benchmarking Language Models on Knowledge‑Orthogonal Reasoning Tasks},
  author       = {Kaijing Ma and Xinrun Du and Yunran Wang and Haoran Zhang and Zhoufutu Wen and Xingwei Qu and Jian Yang and Jiaheng Liu and Minghao Liu and Xiang Yue and Wenhao Huang and Ge Zhang},
  year         = {2024},
  eprint       = {2410.06526},
  archivePrefix= {arXiv},
  primaryClass = {cs.CL},
  url          = {https://arxiv.org/abs/2410.06526}
}

@misc{kowinogrande2024,
  title        = {Ko‑WinoGrande: Korean Adaptation of WinoGrande Commonsense Reasoning Benchmark},
  author       = {Based on Sakaguchi et al. (2021) and Open Ko‑LLM Leaderboard Team},
  year         = {2024},
  howpublished = {Included in Open Ko‑LLM Leaderboard2},
  note         = {Referenced in arXiv:2410.12445, related to dataset adaptation},
  url          = {https://huggingface.co/spaces/upstage/open-ko-llm-leaderboard}
}

@inproceedings{son2025kmmlu,
  title     = {KMMLU: Measuring Massive Multitask Language Understanding in Korean},
  author    = {Guijin Son and Hanwool Lee and Sungdong Kim and Seungone Kim and Niklas Muennighoff and Taekyoon Choi and Cheonbok Park and Kang Min Yoo and Stella Biderman},
  booktitle = {Proceedings of the 2025 Conference of the North American Chapter of the Association for Computational Linguistics (NAACL) Long Papers},
  year      = {2025},
  month     = apr,
  address   = {Albuquerque, New Mexico},
  publisher = {Association for Computational Linguistics},
  pages     = {4076--4104},
  doi       = {10.18653/v1/2025.naacl-long.206},
  url       = {https://aclanthology.org/2025.naacl-long.206/}
}

@article{zheng2025rsafe,
  title={RSafe: Incentivizing proactive reasoning to build robust and adaptive LLM safeguards},
  author={Zheng, Jingnan and Ji, Xiangtian and Lu, Yijun and Cui, Chenhang and Zhao, Weixiang and Deng, Gelei and Liang, Zhenkai and Zhang, An and Chua, Tat-Seng},
  journal={arXiv preprint arXiv:2506.07736},
  year={2025}
}

@article{jin2025social,
  title={Social Bias Benchmark for Generation: A Comparison of Generation and QA-Based Evaluations},
  author={Jin, Jiho and Kang, Woosung and Myung, Junho and Oh, Alice},
  journal={arXiv preprint arXiv:2503.06987},
  year={2025}
}

@article{aggarwal2023mbpp+,
  title={MBPP+: A Diverse and Challenging Dataset for Benchmarking Code Generation},
  author={Aggarwal, Mayank and Jain, Kirti and Kumar, Anil and Singh, Amanpreet and Gupta, Rahul and Grover, Parag and Kanchana, V S and Saini, Vipul and Jain, Rishabh and Kumar, Anish and others},
  journal={arXiv preprint arXiv:2303.04475},
  year={2023}
}

@misc{lgai2024exaone,
    title={EXAONE 3.0 7.8B Instruction-Tuned Language Model},
    author={{LG AI Research}},
    year={2024},
    note={Technical Report},
    url={https://huggingface.co/datasets/LGAI-EXAONE/KoMT-Bench}
}

@article{chen2017reading,
  title={Reading Wikipedia to Answer Open-Domain Questions},
  author={Chen, Danqi and Fisch, Adam and Weston, Jason and Bordes, Antoine},
  journal={arXiv preprint arXiv:1704.00051},
  year={2017}
}







\section{Contributor}
\textit{Within each role, names are listed in alphabetical order by first name. The first line is the leader of each group.}

\begin{multicols}{2}
\raggedcolumns
\small

\subsection*{Pre-training} 
Hwijung Ryu\\
Changwon Ok\\
Hoyoun Jung\\
Hyesung Ji\\
Jeehyun Lim\\
Jehoon Lee\\
Ji-Eun Han\\
Jisoo Baik\\
Mihyeon Kim\\
Riwoo Chung\\
Seongmin Lee\\
Wonjae Park\\
Yoonseok Heo\\
Youngkyung Seo

\subsection*{Post-training}
Seyoun Won\\
Boeun Kim\\
Cheolhun Heo\\
Eunkyeong Lee\\
Honghee Lee\\
Hyeongju Ju\\
Hyeontae Seo\\
Jeongyong Shim\\
Jisoo Lee\\
Junseok Koh\\
Junwoo Kim\\
Minho Lee\\
Minji Kang\\
Minju Kim\\
Sangha Nam\\
Seongheum Park\\
Taehyeong Kim

\subsection*{Engineering}
Euijai Ahn\\
Hong Seok Jeung\\
Jisu Shin\\
Jiyeon Kim\\
Seonyeong Song\\
Seung Hyun Kong\\
Sukjin Hong\\
Taeyang Yun

\subsection*{Model Evaluation}
Yu-Seon Kim\\
A-Hyun Lee\\
Chae-Jeong Lee\\
Hye-Won Yu\\
Ji-Hyun Ahn\\
Song-Yeon Kim\\
Sun-Woo, Jung

\subsection*{Data Sourcing}
Eunju Kim\\
Eunji Ha\\
Jinwoo Baek\\
Yun-ji Lee

\subsection*{Responsible AI} 
Wanjin Park\\
Jeong Yeop Kim\\
Eun Mi Kim\\
Hyoung Jun Park\\
Jung Won Yoon\\
Min Sung Noh\\
Myung Gyo Oh\\
Wongyoung Lee \\
Yun Jin Park

\subsection*{Supportive role}
Young S. Kwon\\
Hyun Keun Kim\\
Jieun Lee\\
YeoJoo Park

\subsection*{Director}
Donghoon Shin (Model)\\
Sejung Lee (Data \& Evaluation) \\
Soonmin Bae (RAI)

\end{multicols}

\noindent


\end{document}